\documentclass[conference]{IEEEtran}
\IEEEoverridecommandlockouts
\usepackage{cite}
\usepackage{amsmath,amssymb,amsfonts}
\usepackage{algorithmic}
\usepackage{graphicx}
\usepackage{textcomp}
\usepackage{xcolor}
\usepackage[numbers]{natbib}
\usepackage[utf8]{inputenc} 
\usepackage[T1]{fontenc}    
\usepackage{hyperref}       
\usepackage{url}            
\usepackage{booktabs}       
\usepackage{amsfonts}       
\usepackage{nicefrac}       
\usepackage{microtype}      
\usepackage{lipsum}		
\usepackage{listings}
\usepackage{xcolor}
\usepackage{booktabs}
\usepackage{caption}
\usepackage{adjustbox}
\usepackage{makecell}
\def\BibTeX{{\rm B\kern-.05em{\sc i\kern-.025em b}\kern-.08em
    T\kern-.1667em\lower.7ex\hbox{E}\kern-.125emX}}
\begin{document}

\title{What Works for 'Lost-in-the-Middle' in LLMs? A Study on GM-Extract and Mitigations\\
}

\author{\IEEEauthorblockN{Mihir Gupte}
\textit{General Motors}\\
mihir.gupte@gm.com
\and
\IEEEauthorblockN{Eshan Dixit}
\textit{General Motors}\\
eshan.dixit@gm.com
\and
\IEEEauthorblockN{Muhammad Tayyab}
\textit{General Motors}\\
muhammad.2.tayyab@gm.com
\and
\IEEEauthorblockN{Arun Adiththan}
\textit{General Motors}\\
arun.adiththan@gm.com
}

\maketitle

\begin{abstract}
\label{sec:abstract}
The diminishing ability of large language models (LLMs) to effectively utilize long-range context-the \textbf{"lost-in-the-middle"} phenomenon-poses a significant challenge in retrieval-based LLM applications. To study the impact of this phenomenon in a real-world application setting, we introduce \textbf{GM-Extract}, a novel benchmark dataset meticulously designed to evaluate LLM performance on retrieval of control variables. To accurately diagnose failure modes, we propose a simple yet elegant evaluation system using two distinct metrics: one for spatial retrieval capability (\textbf{Document Metric}) and the other for semantic retrieval capability (\textbf{Variable Extraction Metric}). We conduct a systematic evaluation of 7-8B parameter models on two multi-document tasks (\textbf{key-value extraction} and \textbf{question-answering}), demonstrating a significant change in retrieval performance simply by altering how the data is represented in the context window.  While a distinct U-shaped curve was not consistently observed, our analysis reveals a clear pattern of performance across models, which we further correlate with perplexity scores. Furthermore, we perform a literature survey of mitigation methods, which we categorize into two distinct approaches: \textbf{black-box} and \textbf{white-box} methods. We then apply these techniques to our benchmark, finding that their efficacy is highly nuanced. Our evaluation highlights scenarios where these strategies successfully improve performance, as well as surprising cases where they lead to a negative impact, providing a comprehensive understanding of their utility in a practical context.
\end{abstract}


\section{Introduction}
\label{sec:introduction}

The proliferation of Large Language Models (LLMs) in retrieval-based applications is largely driven by their extensive context windows\cite{pawar2024and}, which allow for the dynamic injection of large volumes of information. This capability has been leveraged in diverse domains, including agent-assist systems\cite{liuagentbench}, healthcare\cite{bedi2024systematic}, and the automotive industry\cite{wang2025automating}. However, the reliability of these models is challenged by a well-documented limitation in their ability to utilize long contexts effectively. Recent studies have identified a "lost-in-the-middle" phenomenon\cite{liu2024lost}, wherein models exhibit a U-shaped performance curve: they reliably retrieve information from the beginning (primacy bias) and end (recency bias) of the context but struggle significantly with information located in the middle. This positional bias undermines their dependability in critical applications where accurate context retrieval is paramount.

While the "lost-in-the-middle" problem is established, its manifestation on domain-specific, structured data and the efficacy of mitigation techniques in such scenarios remain under-explored. To address this gap, we introduce \textbf{GM-Extract}, a novel benchmark derived from a practical automotive application involving 92 distinct control variables. We evaluate model performance on two downstream tasks: Key-Value Extraction and Question-Answering. Our analysis on previous-generation models like LLaMA-2-7B (non-instruct fine-tuned)\cite{touvron2023llama} and Vicuna-7B-v1.5 (instruct fine-tuned on LLaMa-2-27B)\cite{chiang2023vicuna}, which have a 4096-token context window, reveals that while a distinct U-curve is not always present, significant positional degradation persists. In contrast, a modern model with a 128k token context, LLaMA-3.1-8B\cite{grattafiori2024llama}, achieves near-perfect performance on the same 4k-token task, confirming that performance is a function of the model's context length. However, we demonstrate that this advantage diminishes as the context window is filled, suggesting the problem is one of relative, not absolute, position.

This paper further investigates two critical but often overlooked factors: data representation and the nature of retrieval failure. We find that merely altering the data format-from a structured dictionary (for Key-Value Extraction) to a natural language paragraph (for Question-Answering)-dramatically impacts retrieval accuracy as well as the perplexity of the model.

To dissect these failures, we propose two distinct evaluation metrics: the Document Metric, which measures a model's ability to identify where the correct information is located, and the Variable Extraction Metric, which measures its ability to retrieve what the information is. Our findings indicate that models often fail on the Document Metric even when succeeding on the Variable Extraction Metric, pointing to a lack of spatial awareness within their context. 

\begin{figure*}[h]
    \includegraphics[width=\linewidth]{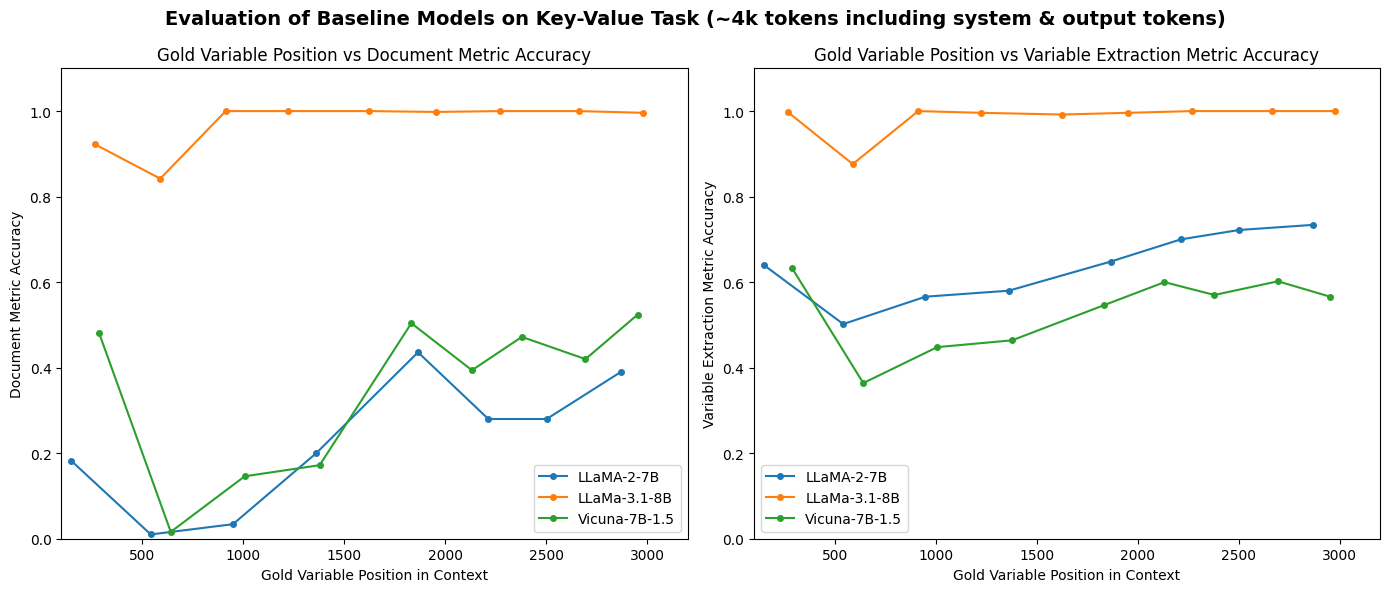}
    \caption{Performance of baseline models on Document Metric of Key-Value Task (left). Performance of baseline models on Variable Metric of Key-Value Task (right)}
    \label{fig:baseline-kv-results}
    \vspace{0.3cm}
    \centering
    \includegraphics[width=\linewidth]{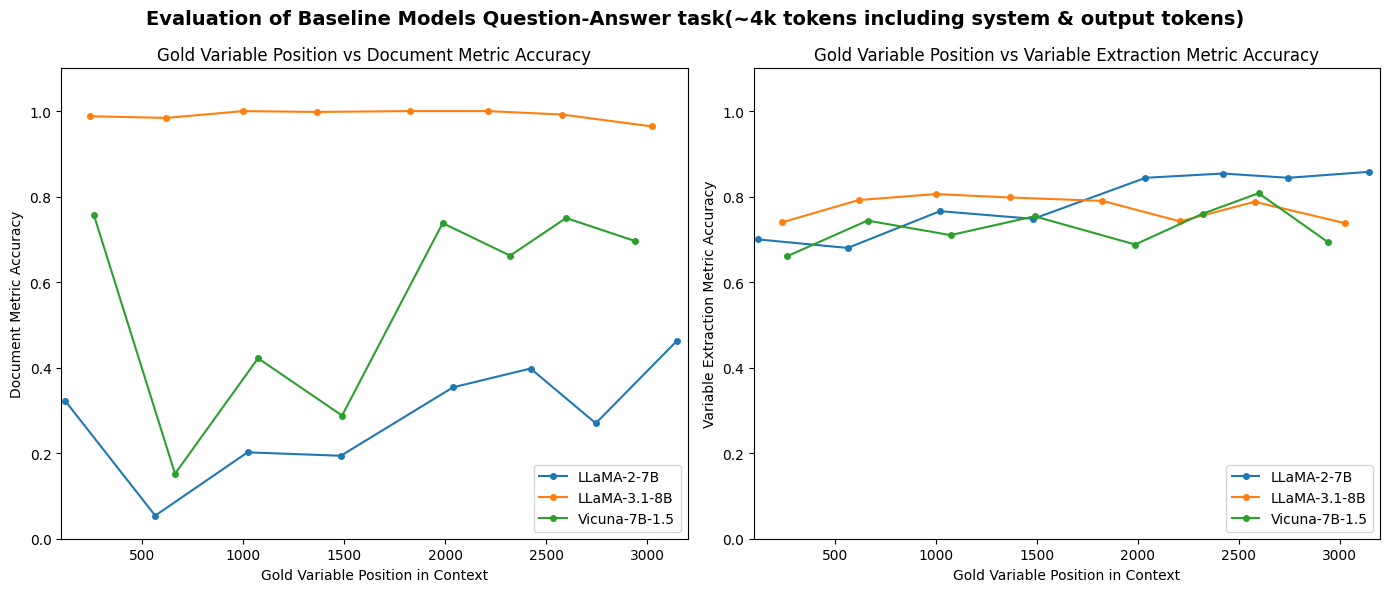}
    \caption{Performance of baseline models on Document Metric of Question-Answer Task (left). Performance of baseline models on Variable Metric of Question-Answer Task (right)}
    \label{fig:baseline-qa-results}
\end{figure*}

We can observe an example of the profound difference in performance caused by changes in data representation in Figure \ref{fig:baseline-kv-results} and Figure \ref{fig:baseline-qa-results} across both metrics. All baseline models\footnote{By Baseline models, we utilize off-the-shelf pre-trained LLMs}, on average, perform significantly better when tasked with retrieving \textit{what} the gold variable is (semantic understanding) compared to \textit{where} it is located (spatial retrieval). By gold variable, we mean the variable which is to be retrieved. Furthermore, the shift in data representation- from the dictionary-like format for Key-Value to the paragraph-like format in Question-Answer- alters performance significantly across all evaluated models. More information on data representation can be found in Section \ref{subsec:data_repr}.

Finally, we systematically evaluate several existing mitigation techniques, which we categorize as black-box and white-box methods, and find their effectiveness to be highly conditional on the model training and data representation.
To summarize, our contributions in this paper are the following:

\begin{enumerate}
    \item We propose \textbf{GM-Extract}, a new benchmark for long-context retrieval based on a practical industrial use case, along with two evaluation tasks and two novel metrics (Document Metric and Variable Extraction Metric) to better diagnose retrieval failures.
    \item We provide an empirical analysis demonstrating that model performance on positional retrieval is highly sensitive to the input data's representation format, a finding we correlate with changes in model perplexity.
    \item We conduct a systematic evaluation of prominent mitigation methods, revealing that their efficacy is not universal and depends significantly on the specific model and task format, highlighting the complexity of resolving the "lost-in-the-middle" problem.
\end{enumerate}
\clearpage
\section{Related Work}
\label{sec:related_work}

\subsection{Long-Context Language Models and the Lost-in-the-Middle Problem}
Large Language Models have demonstrated a significant inability to effectively utilize information within long contexts\cite{liu2024lost, hsieh2024found, zhang2025lost}. This phenomenon, known as the "lost-in-the-middle" problem, manifests as a distinct primacy bias, where models prioritize information at the beginning of the context, and a recency bias, where they prioritize information at the end. The most surprising aspect of this pattern is its pervasiveness across a wide range of models, irrespective of their parameter count or theoretical context length, as highlighted by benchmarks like RULER\cite{hsieh2024ruler}, LongBench\cite{bai2023longbench}, ZeroSCROLLS\cite{shaham2022scrolls}. The RULER benchmark, for instance, has shown that many models struggle with tasks even when utilizing only a fraction of their claimed context window, a pattern that we also observe in our own experiments.

A diverse body of literature has been proposed to mitigate this issue, with much of the work stemming from an understanding of the models' pre-training and their inability to generalize beyond the sequence lengths observed during training\cite{kaiokendev_context_2023, chen2023extending}. If a model is presented with a sequence longer than it was trained on, its performance can collapse significantly. A substantial amount of research, therefore, focuses on techniques to extend this predefined generalizable length.

One of the fundamental components governing sequence length is the positional encoding mechanism. Rotary Positional Embeddings (RoPE)\cite{su2024roformer} is a popular method used in transformer-based models like LLaMa and LLaMa-2. RoPE works by rotating the query and key vectors for each token based on their position, allowing the model to incorporate positional information in a way that respects the relative distance between tokens. The rotation of the vectors is a function of the token's position, and the change in the angle between the vectors of two tokens is directly related to their relative distance, $|a-b|$. A key challenge with RoPE is that this angular distance can cause attention scores to decay as the token distance increases, making it difficult for models to attend to distant tokens.

Prompted by this limitation, several prominent works have proposed methods to modify or replace RoPE. One of the first was Positional Interpolation (PI)\cite{chen2023extending}, which rescales the input positions to fit within the range observed during training. Following this, other methods such as YARN\cite{peng2023yarn} and Ms-PoE\cite{zhang2024found} have been introduced, all focusing on altering how positional information is encoded to artificially increase the effective context length. In this paper, we make use of the Ms-PoE method, with more details provided in Section \ref{subsec:ms-poe}.

Another research direction explores the finding that hidden state channels also convey significant information about the positional states of tokens, a result of the causal attention mask\cite{yu2024mitigate} which showed that attention for positional embeddings decays for the middle channels of the hidden states in transformer models. This suggests that tokens at the beginning and end of the context are more likely to be prioritized. To address this issue, the authors proposes artificially scaling certain dimensions of the hidden states to ensure a more uniform distribution of attention across all positions. We incorporate this approach as one of our mitigation methods, as detailed in Section \ref{subsec:hss}.

Finally, a breadth of work focuses on data-driven solutions \cite{an2024make}. The core idea of this approach is to fine-tune or train the model so that it learns to uniformly distribute attention across its entire context window. This is achieved by modifying the training data such that the outputs necessitate attending to tokens from all parts of the input, especially the middle. While we are constrained by resources, we adopt a method inspired by \cite{an2024make} to create modifications and permutations of our benchmark task. We use Low Rank Adaptation (LoRA)\cite{hu2022lora} as a performance efficient fine-tuning (PEFT) method for tuning our model with these changes. We also evaluate our benchmark on a publicly released model that has been trained with the IN2-Training methodology.

In this work, we explore implmentations of Ms-PoE\cite{zhang2024found}, hidden-state scaling\cite{yu2024mitigate} \& IN2-Training\cite{an2024make}. A detailed rationale for these methods is provided in Section \ref{subsec:overview}.

\subsection{Alternative Positional Encoding and Streaming Context Methods}

Beyond the primary focus on correcting positional decay within the RoPE framework (as explored by PI, YARN, and MS-POE), other notable works have addressed long context through alternative positional encoding paradigms. These include CLEX \cite{chen2023clex}, which leverages context length extrapolation, Self-Extend \cite{jin2024llm}, and the foundational ALiBi (Attention with Linear Biases) \cite{press2021train}, which introduces a non-learned bias to the attention scores proportional to the distance between query and key tokens.

A separate, yet related, body of literature focuses on modifying the attention mechanisms to enable the processing of streaming or non-finite context efficiently. These methods are typically aimed at reducing the computational overhead of the key-value cache, rather than fixing the retrieval bias itself. Noteworthy techniques in this area include creating attention sinks \cite{xiao2023efficient}, which reserve slots for early tokens to prevent catastrophic forgetting; methods like LM-Infinite and attention sorting \cite{peysakhovich2023attention}; and techniques utilizing attention buckets \cite{chen2023fortify} or modifications to the RoPE decay structure, such as LongRoPE \cite{ding2024longrope}. While these methods enhance the model's capacity for long context, our work is primarily focused on the quality of information retrieval within a defined, fixed window.

\section{Methodology}
\label{sec:methods}

\subsection{Motivation Behind Models Used}
In this paper, we select three distinct language models-LLaMa-2-7B\cite{touvron2023llama}, Vicuna-7B-1.5\cite{chiang2023vicuna}, and LLaMa-3.1-8B\cite{grattafiori2024llama}-to conduct our baseline evaluations. All experiments, including the investigation of mitigation methods, are performed on models within the 7-8B parameter range. This deliberate choice serves two primary purposes: first, to demonstrate that our proposed tasks are achievable by models of this scale, thereby showing that task performance is not solely a function of model size. However, it is important to also note that models of larger parameter sizes have been found to demonstrate a similar trend of context utilization\cite{liu2024lost} despite higher performance accuracy; and second, to adhere to institutional resource constraints that limit our experimental setup to a single GPU and a pre-approved set of models. Despite these constraints, the selected models represent a breadth of existing language models and their fine-tuning paradigms.

For example, LLaMa-2-7B is utilized as a representative of a non-instruct-tuned model, allowing us to evaluate the raw performance of a model not explicitly optimized for conversational or instruction-following tasks. In contrast, Vicuna-7B-1.5 is a LLaMa-2-7B model that has been fine-tuned for conversational purposes on user-shared dialogues from ShareGPT. Evaluating our tasks on Vicuna thus provides a fair and direct comparison between the performance of instruct-tuned and non-instruct-tuned models. Both LLaMa-2-7B and Vicuna-7B-1.5 are based on the same architecture with a context window of 4096 tokens.

The LLaMa-3.1-8B model is included to represent a newer generation of instruct-tuned models that boast a significantly larger context window, claiming the ability to attend to up to 128k tokens. Its near-perfect performance on our benchmark with approximately 4k tokens highlights the clear implications of context window size on model performance. A more detailed exploration of its performance degradation as context length increases is provided in Section \ref{subsec:degradation}.

\begin{table}[h!]
    \centering
    \begin{tabular}{|l|c|c|}
    \hline
    \textbf{Model} & \textbf{Context Window Size} & \textbf{Instruct Fine-tuned} \\
    \hline
    LLaMa-2-7B & 4096 tokens & No \\
    \hline
    Vicuna-7B-1.5 & 4096 tokens & Yes \\
    \hline
    LLaMa-3.1-8B & 128k tokens & Yes \\
    \hline
    \end{tabular}
    \caption{Summary of Models for Baseline Evaluation and Their Key Characteristics}
\end{table}

\subsection{Classification of Mitigation Methods}
Most existing mitigation methods for the "lost-in-the-middle" problem can be broadly classified into two categories: white-box approaches and black-box approaches. This distinction is based on whether the method requires internal modifications to the model's architecture or relies solely on external factors.

\begin{enumerate}
\item \textbf{White-box Approaches:} These approaches involve modifying the internal architecture, parameters, or training objectives of a specified model to mitigate context utilization issues. This class of methods includes popular techniques such as positional interpolation\cite{chen2023extending}, Ms-PoE\cite{zhang2024found}, YARN\cite{peng2023yarn}, hidden state scaling\cite{yu2024mitigate}, creation of attention sinks within the transformer architecture\cite{xiao2023efficient}, etc. A significant advantage of many of these methods is that they can be applied at inference-time, thereby avoiding the substantial overhead associated with re-training a large model. In this paper, we experiment with Ms-PoE and hidden state scaling to evaluate their effectiveness.

\item \textbf{Black-box Approaches:} In this category, the model's architecture is treated as a fixed component. These methods seek to address the problem purely by changing factors outside the model, such as the input data. A prominent example of this is the IN2-training approach \cite{an2024make}, which is a purely data-driven solution. Such methods often involve significant resource utilization and training overhead. The underlying principle is that by altering the data distribution seen by the model during fine-tuning, the model will learn to generalize more effectively on long-context tasks. In our work, we adopt a method inspired by this approach, fine-tuning our model on a benchmark similar to the IN2-Training as well as using a publicly available model that has been fine-tuned with IN2-training to evaluate its performance on our benchmark.

\end{enumerate}

\subsection{Overview of Mitigation Methods Used}
\label{subsec:overview}
In this section, we elaborate in detail about the mitigation methods we use for our paper:
\subsubsection{Multi-Scale Positional Encoding (White-box Approach)}
\label{subsec:ms-poe}
Multi-Scale Positional Encoding (Ms-PoE)\cite{zhang2024found} is a plug-and-play white-box approach designed to address the long-context limitations of pre-trained Large Language Models without requiring fine-tuning. The method focuses on modifying how a model's positional embeddings, particularly Rotary Positional Embeddings (RoPE)\cite{su2024roformer}, encode token positions. RoPE, while effective, has a known limitation: the attention score between two tokens decays as the distance between their positions, $|a-b|$, increases. This long-term decay property is a key contributor to the "lost-in-the-middle" problem, as models struggle to attend to tokens far from their current position.

RoPE is a method to encode positional information of the token embeddings. It can be represented by the following formula:
\begin{equation}
    f (\mathbf{q}_m, m)^T f (\mathbf{k}_n, n) = g(\mathbf{q}_m, \mathbf{k}_n, m - n)
    \label{eq:rope_formula}
\end{equation}

where, \textit{q} and \textit{k} represent the query and key embeddings respectively at positions \textit{m} and \textit{n}. 

The function \textit{f} can be broadly represented as:
\begin{equation}
    f(x, m) = xe^{im\theta}
\end{equation}

Finally, to calculate the attention score, RoPE computes the real part of $f (\mathbf{q}_m, m)^T f (\mathbf{k}_n, n)$.
\paragraph{}
Ms-PoE's core innovation is to re-scale the position indices with distinct scaling ratios for different attention heads, effectively creating a multi-scale context fusion from short to long distances. It operates on the principle that not all attention heads are equally sensitive to positional information. Some heads are highly "position-aware" and essential for capturing local syntactic relationships, while others are less so and are better suited for long-range reasoning. Ms-PoE leverages this by applying a smaller scaling factor to the position-aware heads to preserve their learned knowledge and a larger scaling factor to position-unaware heads to relieve the long-term decay effect. 

By applying a rescaling ratio \textit{r}, potential long-term decay effects claim to be avoided. This is done by keeping this ratio \textit{r} within a linear range of $R_{max}$ and $R_{min}$ given by the following formula:
\begin{equation}
    r_i = R_{min} + (i - 1)(R_{max} - R_{min})/(n_h - 1)
\end{equation}
\paragraph{}
This rescaling ratio is applied in the following manner:
\begin{equation}
    f(x, m/r) = xe^{i(m/r)\theta}
\end{equation}

\subsubsection{Hidden State Scaling (White-box Approach)}
\label{subsec:hss}
In this method the authors\cite{yu2024mitigate} modify the positional hidden states of a model to mitigate the lost-in-the-middle problem. The core idea is based on the observation that while values of hidden states are not perfectly monotonic with token position, they are often "roughly monotonic," meaning their values generally increase or decrease with position.

To identify and select the optimal channels for scaling, the method employs a multi-step search algorithm:
\begin{enumerate}
\item \textbf{Identifying Channels:} A polynomial curve is fitted to the hidden state values of each channel to determine if it is "roughly monotonic" across more than a quarter of the model's layers.
\item \textbf{Evaluating Smoothness:} A smoothness score is calculated for each roughly monotonic channel to measure how closely it resembles an ideal monotonic curve. Only the top-K smoothest channels (with a default of K=10) are selected.
\item \textbf{Determining Scale Factor:} The selected channels are evaluated using a small calibration dataset for a retrieval task. A grid search is performed over a set of scale factors ($\{0.5, 0, -0.5, -1\}$) to find the factor that minimizes the loss.
\end{enumerate}
This process selects a single channel and a specific scaling factor that are applied to the positional hidden states during attention calculation, specifically for the final token of the input sequence. This focused approach is based on a series of ablation experiments that found scaling just one channel and modifying only the last token's attention yields the best results.

This can broadly be represented as follows, for a given sequence of length \textit{l}, the query states \textit{$\bar{q_l}$} and key states \textit{K} are modified by a function \textit{P}:
\begin{equation}
    \bar{q_l} = P(W^{Q}f(h(l),p,s),l),\bar{K} = P(W^{K}f(h,p,s),[1,2,...,l]) 
\end{equation}

where the function f represents the \textit{p}-th channel of \textit{h} being scaled by a factor \textit{s}. The combined attention output \textit{z} is calculated by:
\begin{equation}
    z = 
    \begin{cases}
        \text{Softmax}\left(\frac{q_i K^T + \text{Mask}}{\sqrt{d}}\right)V, & i < l \\
        \\
        \text{Softmax}\left(\frac{\bar{q_l} \bar{K}^T}{\sqrt{d}}\right)V, & i = l
    \end{cases}
\end{equation}

\subsubsection{Information-Intensive Training (Black-box Approach)}
The Information-Intensive (IN2) Training approach\cite{an2024make} represents a black-box mitigation strategy, as it aims to improve a model's long-context capabilities without modifying its architecture. The core principle of this method, as described by the original authors, is to explicitly teach the model that crucial information can reside at any position within a long context. This is achieved by generating a large-scale, long-context question-answering dataset where answers are derived from information strategically and randomly placed within the context.

The training data $D = \{L_i, q_i, a_i\}$, is constructed from a general natural language corpus C. For a given raw text $C_i \in C$, a powerful LLM is used to generate a question-answer pair $q_i$, $a_i$ based on a small, randomly extracted segment $s_i$. A long context \textit{$L_i$} is then synthesized by concatenating the necessary segment $s_i$ with other randomly sampled segments. This process is repeated to create two types of training data: one focused on retrieving specific, fine-grained information and another requiring multi-hop reasoning across multiple, distantly-placed segments.

\begin{equation}
    (q_i, a_i) \sim \text{Prompting}(s_i, I_f; \text{LLM}), \quad L_i = \bigoplus\{\text{Shuffle}(s_i, [r_j])\}
\end{equation}

Our approach is inspired by this methodology. We create a fine-tuning dataset based on our own benchmark, leveraging the principle that fine-tuning on specific data can improve a model's decision boundaries for particular tasks. We fine-tune the Vicuna-7B-1.5 model on this dataset using Parameter-Efficient Fine-Tuning (PEFT), specifically LoRA, to adhere to our resource constraints.

To evaluate the full potential of this approach, we also test the performance of a more intensely trained model, FILM-7B, which was trained on the original dataset using a massive computational setup (300 GPU days). By evaluating our benchmark on this model, we can assess the true effectiveness of the Information-Intensive Training method in mitigating the lost-in-the-middle problem, providing a valuable comparison to our resource-constrained fine-tuning efforts.

\subsubsection{Performance Efficient Information Finetuning (Black-box Approach)}

Our black-box approach is inspired by the Information-Intensive Training (IN2) paradigm, adapted to our institutional resource constraints. The goal is to explicitly train the model to attend uniformly across the entire context window.

\textbf{Creating Data for Fine-tuning:} We constructed our fine-tuning dataset using various permutations of the GM-Extract benchmark, carefully excluding the ten gold variables used in the final evaluation set to ensure unbiased assessment of generalization. From the remaining 82 variables, we generated training samples by placing them at 25 different positional indices within the context. The Vicuna-7B-1.5 model, an instruction-tuned model, was ultimately selected for fine-tuning. Initial experimentation with the base, non-instruction-tuned LLaMa-2-7B model did not yield promising results, indicating that a strong initial instruction-following capability is requisite for the effectiveness of this data-driven mitigation strategy. We generated separate training datasets tailored to the format of each downstream task (Key-Value Extraction and Question-Answering) to ensure domain specificity.

\textbf{Parameter-Efficient Fine-Tuning (PEFT):} Due to computational limitations, we utilized Low-Rank Adaptation (LoRA)\cite{hu2022lora} to perform PEFT. We conducted ablation studies by varying the size of the training set (60k, 100k, and 120k permutation samples) and by exploring different values for the LoRA hyperparameters, rank ($r$) and scaling coefficient ($\alpha$), which govern the rank and magnitude of the update matrices, respectively.

\textbf{Non-Uniform Distribution of Data to Mitigate Positional Bias:} Initial evaluations of the fine-tuned models demonstrated a persistent positional bias, with reduced performance observed at the beginning and middle context regions. To explicitly mitigate this, we implemented a data augmentation strategy focusing on positional density. We created a final optimal training set by combining a general 60k sample set (with uniform positional distribution) with an additional 60k sample set exclusively focusing on permutations where the target variable was placed in the challenging middle document positions. This targeted augmentation strategy yielded the most significant improvements in context utilization.

More results of fine-tuning ablations can be found in Appendix \ref{app:fine-tuning}. We use the best fine-tuned model for evaluation of our results.

\subsection{Evaluation Metrics}
As followed by the authors of \cite{liu2024lost, kandpal2023large, mallen2022not}, we also adhere to accuracy to evaluate performance on all models. In addition to that, we propose two different metrics to evaluate different aspects of the retrieval which are - Document Metric (if the model can retrieve the Document ID of the gold variable accurately) \& Variable Extraction Metric (if the model can retrieve the content of the variable itself accurately). These are explained in more broad detail in this section:
\subsubsection{Document Metric}
The Document Metric measures a model's ability to accurately retrieve the ID of the document where the target variable is located. This metric serves as an indicator of whether the model has an understanding of the spatial location of the ground-truth information within the context. 

However, to ensure the integrity of the evaluation, we address cases where models, particularly non-instruct-tuned models like LLaMa-2-7B, generate multiple IDs in their response. To prevent spurious success, the response is considered correct only if the first generated ID matches the ground-truth document ID. All other responses, including those with multiple IDs, are either counted as incorrect or the user is manually prompted to evalute them. Here is an example:
\begin{figure}[h] 
    \centering
    \includegraphics[width=1\linewidth]{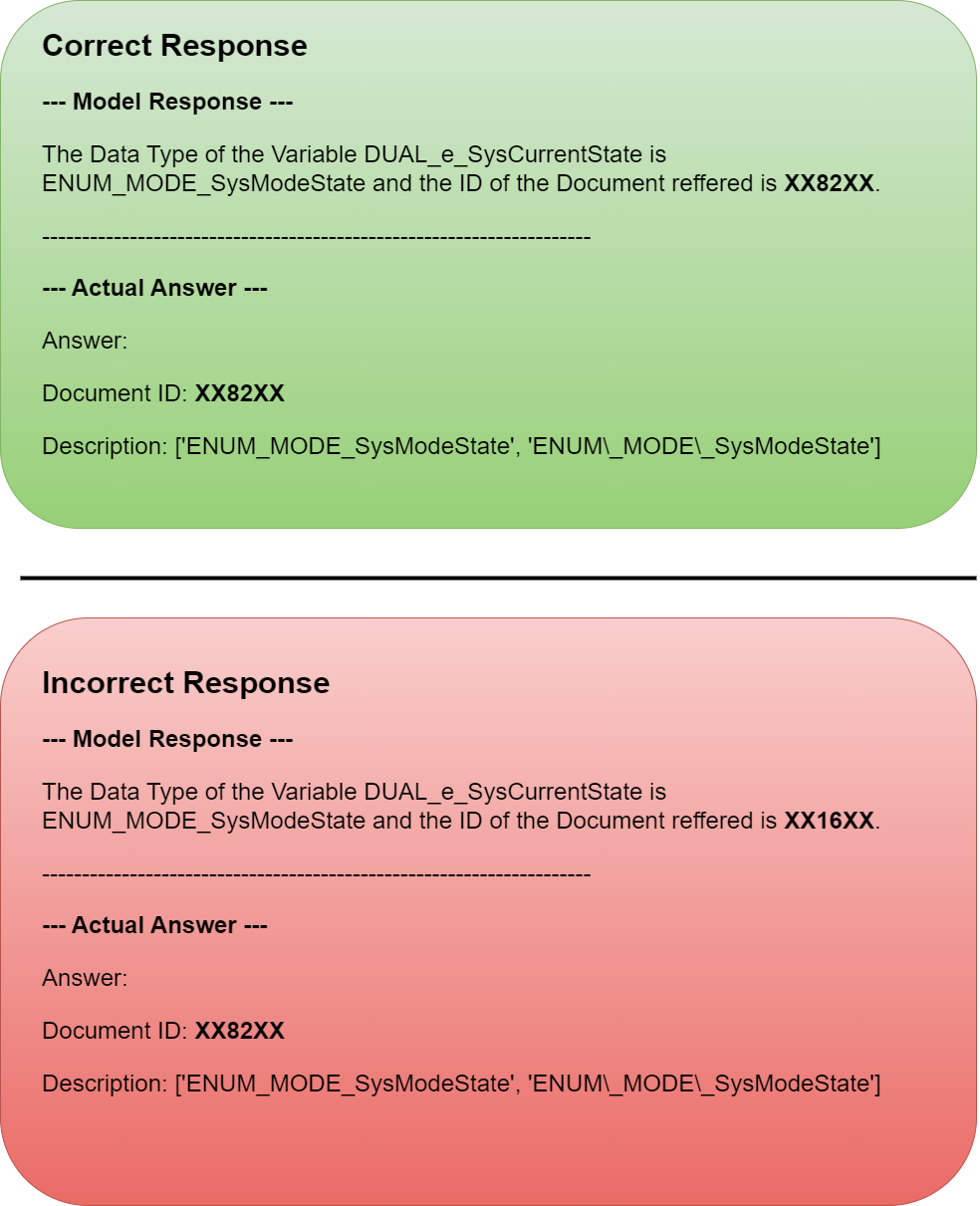} 
    \caption{An Example of Document Metric Evaluation which evaluates spatial retrieval capability of the model }
    \label{fig:document-metric-evaluation}
\end{figure}

The metric is formally defined using an indicator function, $\mathbb{I}$, as follows:
\begin{equation}
    M_{\text{doc}} = \frac{1}{N} \sum_{i=1}^{N} \mathbb{I}(T_i \in P_i)
    \label{eq:doc-metric}
\end{equation}
where $N$ is the total number of samples, $P_i$ is the set of document IDs generated in the model's response for the $i$-th sample, and $T_i$ is the ground-truth document ID.

\subsubsection{Variable Extraction Metric}
The Variable Extraction Metric measures a model's ability to accurately retrieve the content of a target variable. This can be thought of as assessing whether the model understands what the correct answer is, rather than just where it's located. The metric is evaluated as correct if the gold variable's content is present anywhere in the model's generated response.
\begin{figure}[h] 
    \centering
    \includegraphics[width=1\linewidth]{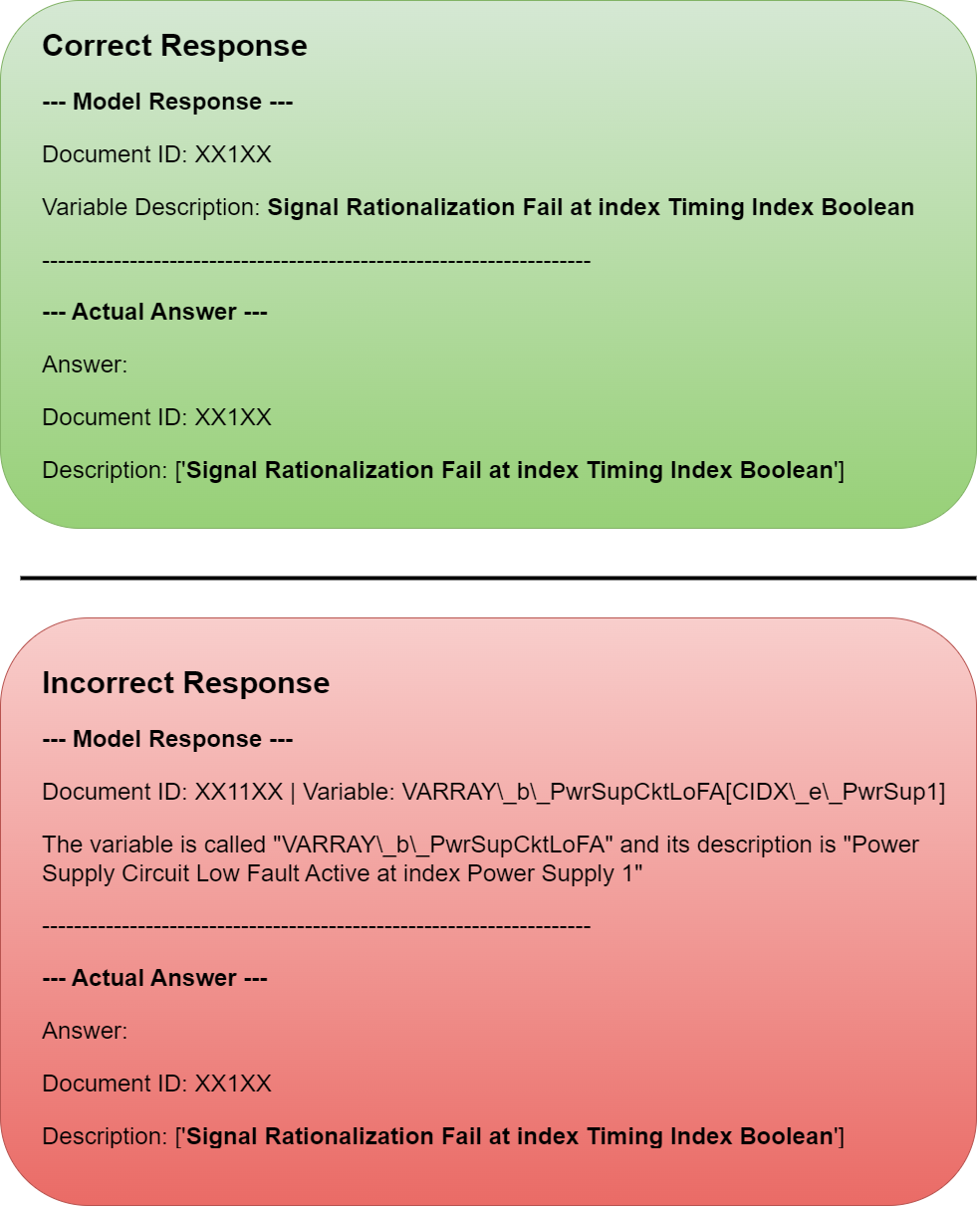} 
    \caption{An Example of Variable Extraction Metric Evaluation which evaluates the semantic retrieval capability of the model }
    \label{fig:variable-metric-evaluation}
\end{figure}

The metric is formally defined using an indicator function, $\mathbb{I}$, as follows:
\begin{equation}
    M_{\text{var}} = \frac{1}{N} \sum_{i=1}^{N} \mathbb{I}(V_i \in P_i)
    \label{eq:var-metric}
\end{equation}
    
where $N$ is the total number of samples, $V_i$ is the ground-truth variable content for the i-th sample, and $P_i$ is the set of words in the model's response for that sample.

\subsubsection{Perplexity}
Perplexity (PP) serves as a quantitative measure of a language model's uncertainty on a given text. A lower perplexity score indicates that the model is more "certain" and assigns a higher probability to the text, suggesting a better fit. Conversely, a high perplexity score signifies a less confident model. In our experiments, we use perplexity to correlate with the observed performance trends, investigating whether the models' inability to utilize middle context corresponds to an increase in their perplexity.

Formally, perplexity is defined as the exponential of the negative average log-likelihood of a sequence of tokens, $W = w_1 w_2 ... w_N$:
\begin{equation}
    PP(W) = \exp\left(-\frac{1}{N} \sum_{i=1}^N \log P(w_i | w_1...w_{i-1})\right)
    \label{eq:perplexity}
\end{equation}
where $N$ is the total number of tokens in the sequence, and $P(w_i | w_1...w_{i-1})$ is the probability of the $i$-th token given the preceding tokens. This probability is computed by the language model.

\section{Experiment \& Benchmark Setup}
\label{sec:data_experiment}

\subsection{GM-Extract: Overview of Benchmark Data}
The \textbf{GM-Extract} benchmark is a novel dataset comprising 92 unique control variables derived from a real-world application setting. Each variable within the dataset is an object containing seven distinct metadata fields, where the core relationship is defined by the unique variable identifier (\texttt{Variable}) and its natural language description (\texttt{Input}). The remaining fields (\texttt{Data Type}, \texttt{Meta-data}, \texttt{SET}, \texttt{ASSERT}, etc.) provide supplementary information on usage, type, and system interaction.

The following structure illustrates a representative sample from the benchmark:
\begin{verbatim}
{
    "Input": "Propulsion System Active 
    Response",
    "Solution IO Type": "Input",
    "Variable": "VAR_PropSysActive_rsp",
    "Data Type": "BOOL",
    "SET": "VAR_PropSysActive_rsp = 
    value_to_set",
    "ASSERT": 
    "CHECK_TRUE(VAR_PropSysActive_rsp)",
    "Meta-data": "The variable 
    VAR_PropSysActive_rsp is of storage 
    type: Variable, data type: 
    Element - single piece of data, 
    and measures the quantity: None"
}
\end{verbatim}

\subsubsection{Data Representation for Downstream Tasks}
\label{subsec:data_repr}
To rigorously test the impact of data formatting on model performance-a key objective of this study-we convert the raw variable metadata into two distinct representations for the downstream tasks. The difference in structure is designed to isolate the models' bias towards structured versus natural language formats.

\paragraph{Key-Value (KV) Formatting}
For the KV extraction task, the data is formatted into a highly structured, machine-readable key-value list. This format is intended to favor models that excel at structured parsing.
\begin{verbatim}
Document XX1XX:
Variable: VAR_PropSysActive_rsp 
Variable Description: Propulsion System 
Active Response
Data Type: BOOL
Metadata: The variable 
VAR_PropSysActive_rsp is of 
storage type: Variable, 
data type: Element - single piece of data, 
and measures the quantity: None
SET Command: `VAR_PropSysActive_rsp 
= value_to_set` 
ASSERT Command: 
`CHECK_TRUE(VAR_PropSysActive_rsp)`
\end{verbatim}

\paragraph{Question-Answering (QA) Formatting}
For the QA task, the same information is reframed into a contiguous, natural language paragraph. This format requires the model to perform comprehension and inference over prose, testing its ability to handle less structured input.
\begin{verbatim}
Document XX1XX:
Variable VAR_PropSysActive_rsp. 
The variable refers to input 
Propulsion System Active Response. 
The variable VAR_PropSysActive_rsp is 
of storage type: Variable, data type: 
Element - single piece of data, 
and measures the quantity: None. 
The Solution IO Type for this v
ariable is Input. The System Data 
Type of this variable is BOOL. 
To set this variable use: 
`VAR_PropSysActive_rsp = value_to_set` 
and to assert this variable use: 
`CHECK_TRUE(VAR_PropSysActive_rsp)`.
\end{verbatim}
The full list of the set of questions evaluated are available in Appendix \ref{app:list-questions}.

\subsection{Experiment Setup}
Our experimental design follows a methodology similar to the widely recognized \textbf{Needle-In-A-Haystack} paradigm \cite{kamradt2023needle, hsieh2024ruler, mohtashami2023random}. The task is straightforward: given $k$ documents placed within a model's context window, we embed a "gold" variable-the target information to be retrieved-within one of these documents. The documents themselves contain information about variables from our benchmark, formatted specifically for each task (e.g., dictionary format for the Key-Value task and natural language paragraphs for the Question-Answering task). Both tasks require the model to identify the relevant information and provide the correct response. We judge performance using our two metrics: the \textbf{Document Metric} and the \textbf{Variable Extraction Metric}. We also measure the perplexity of the model's response and correlate it with the position of the gold variable. This setup serves as a strong demonstration of a model's practical efficacy in a real-world, information-retrieval application.

To thoroughly evaluate the impact of positional bias, we systematically rotate the gold variable across every position within the context window. For the Key-Value task, we use 25 documents, and for the Question-Answering task, we use 22 documents. This configuration results in approximately 4k tokens, fitting within the context window of the models we are evaluating. We measure performance at every third document position to capture a representative sample of positional effects.

To account for the non-deterministic nature of these models, we evaluate each unique experimental configuration 50 times to ensure a statistically robust estimate of the accuracy. For example, for a given question and a specific document position, we run the retrieval task 50 times on the same input, with only the gold variable's position varying from one test to the next. The entire context is kept constant, ensuring that the only variable being tested is the position of the gold information. We also utilize a fixed set of 10 questions to ensure consistency across all evaluations (more details in Appendix \ref{app:list-questions}).

\subsection{Implementation Details}
All experiments, including inference and the implementation of mitigation methods, were executed using the PyTorch and HuggingFace Transformers frameworks on a single NVIDIA A100 80GB GPU. We leveraged LoRA for all Parameter Efficient Fine-Tuning (PEFT) tasks.

\paragraph{Baseline Models}
Our baseline evaluation utilizes two distinct 7B-parameter models: the foundational, non-instruct-tuned \textbf{LLaMa-2-7B} and the instruction-tuned \textbf{Vicuna-7B-1.5}. Both models possess a native 4,096-token context window. For comparison against next-generation models, we also evaluate the \textbf{LLaMa-3.1-8B} model, which features a substantially larger 128k-token context window.

\paragraph{White-Box Mitigation Methods}
For our white-box analysis, we implemented and evaluated Multi-Scale Positional Encoding (\textbf{MS-PoE})\cite{zhang2024found} and Hidden State Scaling\cite{yu2024mitigate} on both the Vicuna-7B-1.5 and LLaMa-2-7B models, strictly following the methodologies outlined in their respective source literatures.

\paragraph{Black-Box Mitigation Methods}
Our primary black-box investigation involved applying Parameter Efficient Fine-Tuning (PEFT) with our Information-Intensive Training strategy to both the Vicuna-7B-1.5 and LLaMa-2-7B models. Due to substantial resource constraints, full fine-tuning was not performed. To evaluate the upper bound of this approach, we instead benchmarked the fully-trained FILM-7B model\cite{an2024make}-which was intensely trained using the IN2 method. It is important to note that FILM-7B utilizes a 32k-token context window; its performance degradation with increasing context length is further explored in Section \ref{subsec:degradation}.

\subsection{Results}

The evaluation results for our models and their respective mitigation strategies are presented in Table \ref{table:doc-metric-performance} (Document Metric) and Table \ref{table:var-metric-performance} (Variable Extraction Metric). Our analysis of these tables leads to several critical observations:

\begin{table*}[htbp]
    \centering
    \caption*{\textbf{Performance of Models Across Tasks on Document Metric}}
    \begin{adjustbox}{width=\textwidth}
      \begin{tabular}{l|ccccccc}
      \toprule
      Models\\  +Mitigations & \multicolumn{7}{c}{\makecell{Document Position \\ \textbf{(KV Task)}}} \\
      \cmidrule(lr){2-8}
          & 1st & 4th & 7th & 10th & 16th & 22nd & Avg \\
      \midrule
      LLaMa-2-7B \\ Baseline       & 0.18 & \textbf{0.01} & \textbf{0.03} & \textbf{0.44} & \textbf{0.28} & \textbf{0.39} & \textbf{0.24}\\
      LLaMa-2-7B \\+ ms-poe       & \textbf{0.50} & 0.00 & 0.01 & 0.29 & 0.17 & 0.32 & 0.21\\
      LLaMa-2-7B \\+ hidden state scaling & 0.17 & 0.01 & 0.03 & 0.16 & 0.14 & 0.18 & 0.14\\
      LLaMa-2-7B \\+ fine-tuning     & 0.04 & 0.00 & 0.00 & 0.00 & 0.02 & 0.08 & 0.02\\
      \hline
      Vicuna-7B \\Baseline        & 0.48 & 0.02 & 0.15 & 0.17 & 0.40 & 0.52 & 0.35\\
      Vicuna-7B \\+ ms-poe        & 0.58 & 0.18 & \textbf{0.43} & \textbf{0.55} & \textbf{0.80} & \textbf{0.58} & 0.46\\
      Vicuna-7B \\+ hidden state scaling & \textbf{0.67} & \textbf{0.22} & 0.33 & 0.32 & 0.66 & 0.50 & \textbf{0.47}\\
      Vicuna-7B \\+ fine-tuning     & 0.45 & 0.06 & 0.29 & 0.48 & 0.57 & 0.65 & 0.43\\
      \hline
      FILM-7B \\(32k context)      & 1.00 & 0.85 & 0.84 & 0.92 & 0.84 & 0.53 & 0.83\\
      \hline
      LLaMa-3.1-8B \\(128k context)   & 0.92 & 0.84 & 1.00 & 1.00 & 1.00 & 0.80 & 0.96\\
      \bottomrule
      \end{tabular}
      
    \begin{tabular}{|ccccccc}
      \toprule
      \multicolumn{7}{c}{\makecell{\\Document Position \\ \textbf{(QA Task)}}} \\
      \cmidrule(lr){1-7}
          1st & 4th & 7th & 10th & 16th & 22nd & Avg \\
      \midrule
      \\
      0.32 & 0.05 & 0.20 & 0.20 & \textbf{0.40} & 0.46 & 0.28\\
      \\
      0.53 & 0.06 & 0.14 & \textbf{0.55} & 0.26 & \textbf{0.52} & \textbf{0.31}\\
      \\
      0.27 & \textbf{0.14} & \textbf{0.23} & 0.15 & 0.32 & 0.29 & 0.23\\
      \\
      \textbf{0.56} & 0.00 & 0.02 & 0.18 & 0.21 & 0.06 & 0.12\\
      \hline
      \\
      \textbf{0.83} & 0.16 & 0.42 & 0.39 & 0.75 & 0.81 & 0.63\\
      \\
      0.80 & 0.22 & \textbf{0.47} & 0.67 & 0.53 & 0.74 & 0.61\\
      \\
      0.74 & 0.18 & 0.36 & 0.20 & 0.80 & 0.77 & 0.59\\
      \\
      0.69 & \textbf{0.30} & 0.26 & \textbf{0.86} & \textbf{0.86} & \textbf{0.86} & \textbf{0.70}\\
      \hline
      \\
      0.89 & 0.85 & 0.91 & 0.91 & 0.91 & 0.73 & 0.87\\
      \hline
      \\
      0.99 & 0.98 & 1.00 & 0.99 & 0.96 & 0.97 & 0.99\\
      \bottomrule
      \end{tabular}
  \end{adjustbox}
      
    \caption{Document Metric Evaluation: \textit{"where is the gold variable located?"}, \textbf{Left:} Performance of Models on Key-Value Task, \textbf{Right:} Performance of Models on Question-Answer. LLaMa-2-7B and Vicuna-7B are 4k context models}
    \label{table:doc-metric-performance}
\end{table*}

\begin{table*}[htbp]
  \centering
  \caption*{\textbf{Performance of Models Across Tasks on Variable Extraction Metric}}
  \begin{adjustbox}{width=\textwidth}
    \begin{tabular}{l|ccccccc}
      \toprule
      Models\\  +Mitigations & \multicolumn{7}{c}{\makecell{Document Position \\ \textbf{(KV Task)}}} \\
      \cmidrule(lr){2-8}
          & 1st & 4th & 7th & 10th & 16th & 22nd & Avg \\
          \midrule
      LLaMa-2-7B \\Baseline       & 0.64 & \textbf{0.50} & \textbf{0.57} & \textbf{0.58} & \textbf{0.70} & \textbf{0.73} & \textbf{0.64}\\
      LLaMa-2-7B \\+ ms-poe       & \textbf{0.67} & 0.45 & 0.50 & 0.47 & 0.57 & 0.58 & 0.55\\
      LLaMa-2-7B \\+ hidden state scaling & 0.44 & 0.39 & 0.51 & 0.49 & 0.64 & 0.61 & 0.54\\
      LLaMa-2-7B \\+ fine-tuning     & 0.44 & 0.42 & 0.40 & 0.46 & 0.62 & 0.40 & 0.47\\
      \hline
      Vicuna-7B \\Baseline        & 0.63 & 0.36 & 0.45 & 0.46 & 0.60 & 0.57 & 0.53\\
      Vicuna-7B \\+ ms-poe        & 0.64 & 0.47 & 0.55 & 0.60 & \textbf{0.80} & 0.60 & 0.60\\
      Vicuna-7B \\+ hidden state scaling & \textbf{0.73} & \textbf{0.56} & \textbf{0.66} & \textbf{0.71} & 0.72 & 0.64 & \textbf{0.65}\\
      Vicuna-7B \\+ fine-tuning     & 0.50 & 0.38 & 0.47 & 0.48 & 0.68 & \textbf{0.72} & 0.59\\
      \hline
      FILM-7B \\(32k context)      & 0.95 & 0.90 & 0.86 & 0.92 & 0.84 & 0.86 & 0.88\\
      \hline
      LLaMa-3.1-8B \\(128k context)   & 0.99 & 0.88 & 1.00 & 0.99 & 1.00 & 0.87 & 0.98\\
      \bottomrule
      \end{tabular}
    \begin{tabular}{|ccccccc}
    \toprule
    \multicolumn{7}{c}{\makecell{\\Document Position \\ \textbf{(QA Task)}}} \\
    \cmidrule(lr){1-7}
    1st & 4th & 7th & 10th & 16th & 22nd & Avg \\
        \midrule
      \\
    \textbf{0.70} & 0.68 & \textbf{0.77} & \textbf{0.75} & \textbf{0.85} & \textbf{0.86} & \textbf{0.79}\\
    \\
    0.66 & \textbf{0.72} & 0.72 & 0.69 & 0.75 & 0.82 & 0.74\\
    \\
    0.59 & 0.59 & 0.72 & 0.73 & 0.73 & 0.81 & 0.71\\
    \\
    0.38 & 0.39 & 0.47 & 0.25 & 0.71 & 0.74 & 0.54\\
    \hline
    \\
    0.64 & 0.70 & 0.75 & 0.77 & 0.73 & 0.70 & 0.72\\
    \\
    0.70 & 0.66 & 0.78 & 0.75 & 0.83 & 0.80 & 0.77\\
    \\
    0.58 & 0.66 & 0.73 & 0.75 & 0.76 & 0.67 & 0.71\\
    \\
    \textbf{0.90} & \textbf{0.86} & \textbf{0.88} & \textbf{0.85} & \textbf{0.89} & \textbf{0.83} & \textbf{0.87}\\
    \hline
    \\
    0.95 & 0.90 & 0.86 & 0.92 & 0.84 & 0.86 & 0.87\\
    \hline
    \\
    0.99 & 0.88 & 1.00 & 0.99 & 1.00 & 0.87 & 0.76\\
    \bottomrule
    \end{tabular}
  \end{adjustbox}
  \caption{Variable Metric Evaluation: \textit{"what is the gold variable?"}, \textbf{Left:} Performance of Models on Key-Value Task, \textbf{Right:} Performance of Models on Question-Answer. LLaMa-2-7B and Vicuna-7B are 4k context models}
  \label{table:var-metric-performance}
\end{table*}

\paragraph{Impact of Data Representation} We observe a significant and consistent change in the Baseline performance simply by altering the data's representation (structured Key-Value vs. Question-Answer format), even when the underlying data content and order remain the same. For the Document Metric, a 15-20\% increase in average scores is observed across all baselines and mitigation methods when using the Question-Answer format. This suggests the model's performance is highly sensitive to the prompt structure, a finding we explore further in relation to perplexity in Section \ref{subsec:data_rep}.

\paragraph{Mitigation Efficacy on Instruct-Tuned Model (Vicuna-7B)} The Key-Value (KV) task sees the most substantial improvement (10-12\% on average) using white-box mitigation methods. Conversely, the Question-Answer (QA) task benefits most from black-box methods (12\% on average) across both metrics. It is noteworthy that for the Document Metric in QA tasks, the average performance of the mitigation methods is slightly worse than the baseline, while other metrics show a 5-7\% improvement.

\begin{figure*}[h] 
    \centering
    \includegraphics[width=1\linewidth]{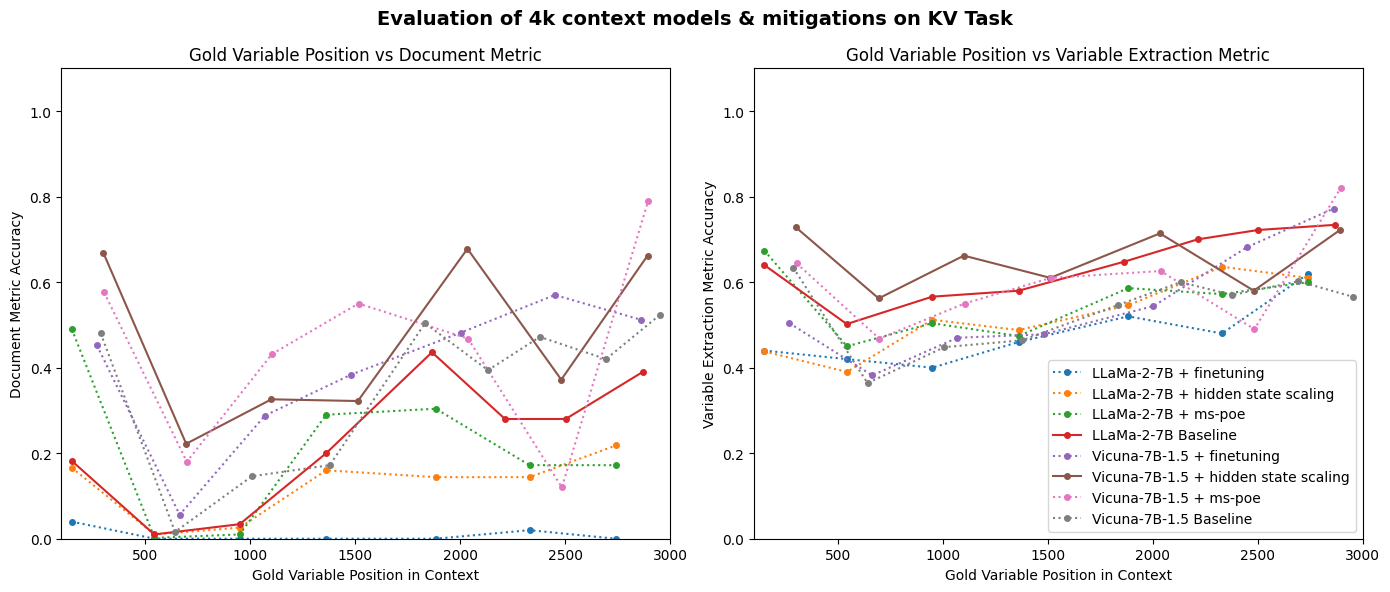} 
    \caption{Performance of LLaMa-2-7B and Vicuna-7B across all mitigations for KV Task}
    \label{fig:kv-mitigation}
\end{figure*}

\begin{figure*}[h] 
    \centering
    \includegraphics[width=1\linewidth]{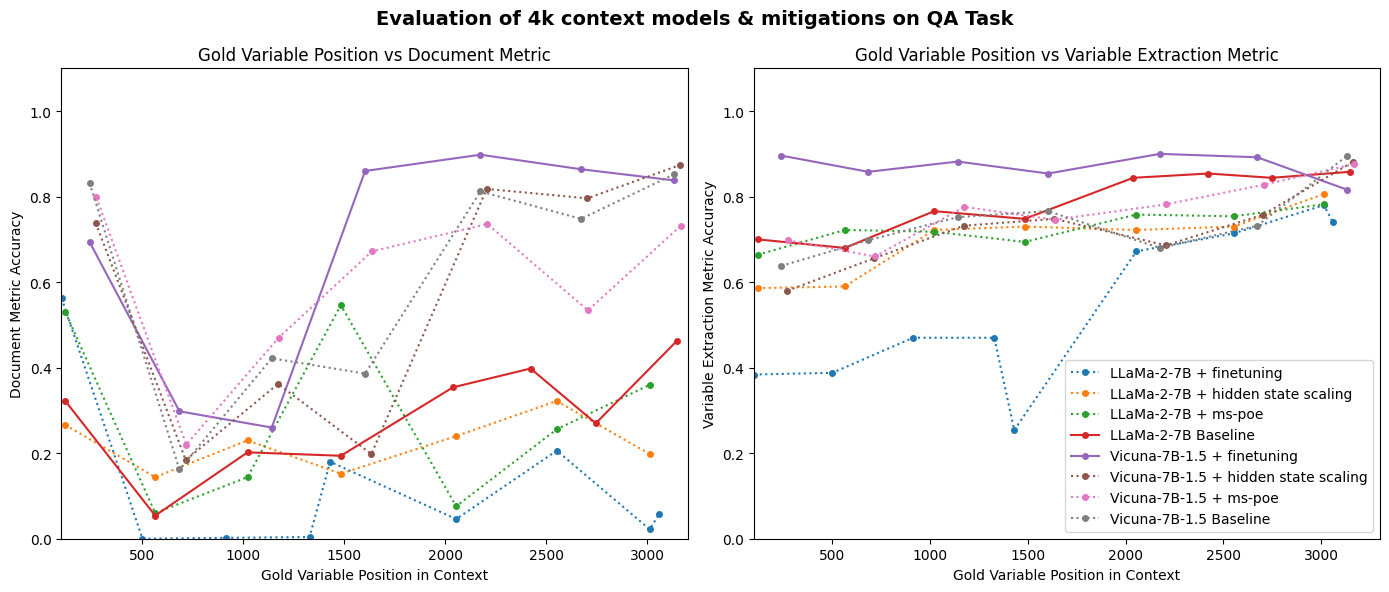} 
    \caption{Performance of LLaMa-2-7B and Vicuna-7B across all mitigations for QA Task}
    \label{fig:qa-mitigation}
\end{figure*}

\paragraph{Mitigation Degradation on Non-Instruct Model (LLaMA-2-7B)} For the non-instruct model, mitigation methods often perform significantly worse than the baseline in almost all scenarios. The only exception is the Document Metric in the QA task, where intermittent performance gains are observed. We hypothesize that the mitigation techniques introduce a formatting bias that negatively interacts with the base model's unaligned training, which we correlate with perplexity in Section \ref{subsec:data_rep}.

\begin{table*}[htbp]
    \centering
    \caption*{\textbf{Performance of Long-Context Models Across Tasks on Document Metric}}
    \begin{adjustbox}{width=\textwidth}
      \begin{tabular}{l|ccccccc}
        \toprule
        Models\\  +Tokens Used & \multicolumn{7}{c}{\makecell{Document Position \\ \textbf{(KV Task)}}} \\
        \cmidrule(lr){2-8}
            & 1st & 13th & 25th & 50th & 75th & 90th & Avg \\
            \midrule
        LLaMa-3.1-8B \\ w/ 4k tokens       & 0.92 & 1.00 & 0.80 & - & - & - & \textbf{0.96}\\
        LLaMa-3.1-8B \\ w/ 12k tokens      & 0.82 & 0.92 & 0.85 & 0.81 & 0.84 & 0.78 & 0.84\\
        \hline
        FILM-7B \\ w/ 4k tokens      & 1.00 & 0.77 & 0.82 & - & - & - & \textbf{0.83}\\
        FILM-7B \\ w/ 12k tokens     & 1.00 & 0.58 & 0.42 & 0.44 & 0.37 & 0.52 & 0.52\\
        \bottomrule
        \end{tabular}
      \begin{tabular}{|ccccccc}
      \toprule
      \multicolumn{7}{c}{\makecell{\\Document Position \\ \textbf{(QA Task)}}} \\
      \cmidrule(lr){1-7}
      1st & 13th & 25th & 50th & 75th & 90th & Avg \\
          \midrule
        \\
      0.99 & 0.99 & 0.98 & - & - & - & \textbf{0.99}\\
      \\
      0.93 & 0.99 & 0.92 & 0.88 & 0.80 & 0.86 & 0.89\\
      \hline
      \\
      0.89 & 0.91 & 0.73 & - & - & - & \textbf{0.87}\\
      \\
      0.99 & 0.86 & 0.86 & 0.60 & 0.57 & 0.55 & 0.77\\
      \bottomrule
      \end{tabular}
    \end{adjustbox}
    \vspace{0.3cm}
    \caption*{\textbf{Performance of Long-Context Models Across Tasks on Variable Extraction Metric}}
    \begin{adjustbox}{width=\textwidth}
      \begin{tabular}{l|ccccccc}
        \toprule
        Models\\  +Tokens Used & \multicolumn{7}{c}{\makecell{Document Position \\ \textbf{(KV Task)}}} \\
        \cmidrule(lr){2-8}
            & 1st & 13th & 25th & 50th & 75th & 90th & Avg \\
            \midrule
        LLaMa-3.1-8B \\ w/ 4k tokens       & 0.99 & 1.00 & 1.00 & - & - & - & \textbf{0.98}\\
        LLaMa-3.1-8B \\ w/ 12k tokens      & 0.94 & 1.00 & 1.00 & 0.99 & 0.95 & 0.96 & 0.97\\
        \hline
        FILM-7B \\ w/ 4k tokens      & 0.95 & 0.80 & 0.86 & - & - & - & \textbf{0.88}\\
        FILM-7B \\ w/ 12k tokens     & 0.96 & 0.76 & 0.79 & 0.78 & 0.81 & 0.83 & 0.80\\
        \bottomrule
        \end{tabular}
      \begin{tabular}{|ccccccc}
      \toprule
      \multicolumn{7}{c}{\makecell{\\Document Position \\ \textbf{(QA Task)}}} \\
      \cmidrule(lr){1-7}
      1st & 13th & 25th & 50th & 75th & 90th & Avg \\
          \midrule
        \\
      0.74 & 0.79 & 0.71 & - & - & - & \textbf{0.76}\\
      \\
      0.87 & 0.75 & 0.67 & 0.75 & 0.73 & 0.72 & 0.73\\
      \hline
      \\
      0.75 & 0.86 & 0.90 & - & - & - & \textbf{0.87}\\
      \\
      0.71 & 0.85 & 0.83 & 0.89 & 0.79 & 0.82 & 0.82\\
      \bottomrule
      \end{tabular}
    \end{adjustbox}
    \caption{Metric Evaluation on Long Context Models, \textbf{Left:} Performance of Models on Key-Value Task, \textbf{Right:} Performance of on Question-Answer Task}
    \label{table:long-context-performance}
\end{table*}

\paragraph{Superiority of Long-Context Models} Models with significantly extended context windows demonstrate near-perfect performance. FILM-7B (32k context) scores in the 83-87\% range, and LLaMA-3.1-8B (128k context) scores in the 96-99\% range across both tasks and the Document Metric. The only notable exception is the Variable Extraction Metric on the QA task for LLaMA-3.1-8B, where the score dips to 76\% on average. This indicates that generalized long-context training is the most optimal solution for alleviating the immediate issue. However, we also observe a degradation of performance in these models when the total context length exceeds 10k, as detailed in Section \ref{subsec:degradation}.

\paragraph{Fundamental Limitation} The maximum performance increase achieved by all mitigation methods is modest (7-12\% for Document Metric and 12-15\% for Variable Extraction Metric). This suggests that even the best methods only alleviate the "lost-in-the-middle" problem, indicating that this limitation is deeply fundamental to how attention is distributed within transformer-based architectures.

\begin{figure*}[h] 
    \centering
    \includegraphics[width=1\linewidth]{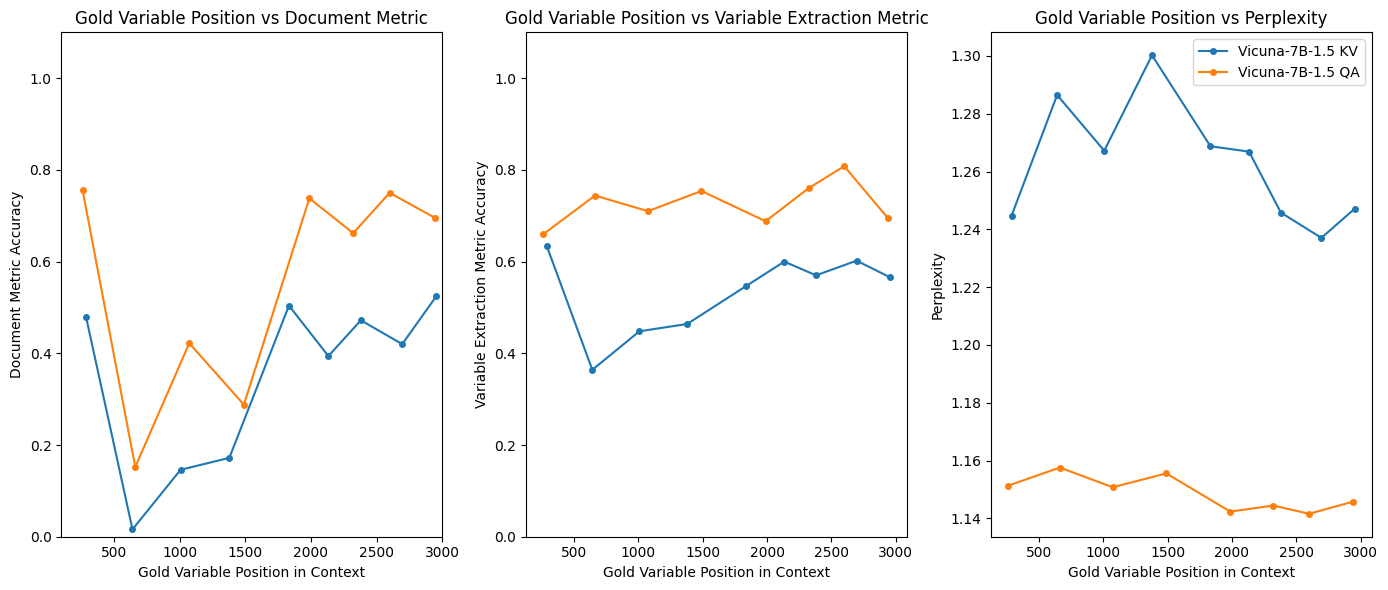} 
    \caption{Performance of Vicuna-7B-1.5 across Tasks}
    \label{fig:vicuna-across-tasks}
\end{figure*}

\begin{figure*}[h] 
    \centering
    \includegraphics[width=1\linewidth]{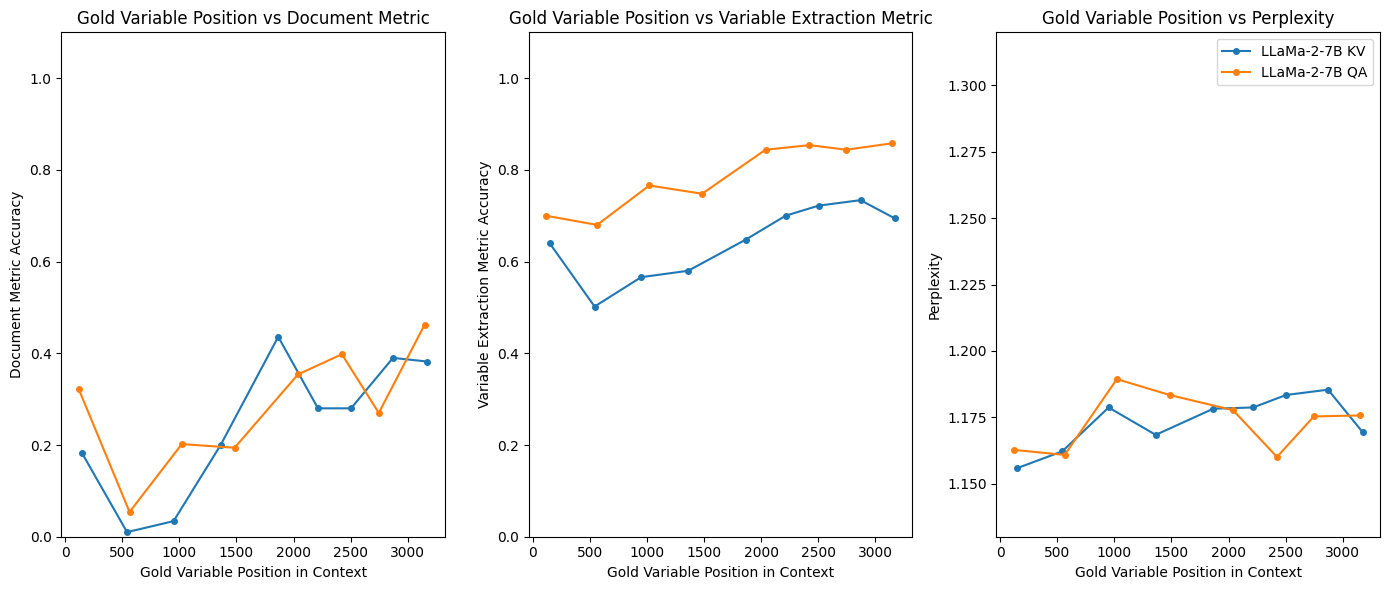} 
    \caption{Performance of LLaMa-2-7B across Tasks}
    \label{fig:llama2-across-tasks}
\end{figure*}

\section{Discussion}
\label{sec:results}

\subsection{Impact of Data Representation on Performance}
\label{subsec:data_rep}
We observe a significant disparity in the retrieval scores for both the Key-Value (KV) and Question-Answering (QA) tasks on the same model. Given that the only manipulation between these two evaluations is the representation of the source data-structured dictionary for KV versus natural language paragraph for QA-this difference suggests a pre-existing internal bias within the models for specific data formats.

This bias is quantitatively supported by the correlation with Perplexity (PP). As shown in Figure \ref{fig:vicuna-across-tasks}, the perplexity scores change markedly not only with the gold variable's position but also with the task format. A model exhibits higher epistemic uncertainty (higher PP) when faced with a data representation that results in lower performance. Our hypothesis is that Language Models are heavily biased toward the specific data representations prevalent during their pre-training and instruction-tuning stages.

For instance, the instruction-tuned Vicuna-7B-1.5 model demonstrates clear performance divergence across formats, as represented in Figure \ref{fig:vicuna-across-tasks}. In contrast, while the effect is less profound for the non-instruct-tuned LLaMa-2-7B base model, a measurable difference in the Variable Extraction Metric is still observed across tasks (Figure \ref{fig:llama2-across-tasks}), underscoring that data representation fundamentally alters a model's certainty and retrieval capability.

\subsection{Superiority of Long-Context Models on Shorter Contexts}
The most direct approach to mitigating the "lost-in-the-middle" problem for sequences up to 4k tokens is by utilizing models pre-trained on significantly larger context windows. The LLaMa-3.1-8B model, which was trained on context sequences up to [insert fact about the length of context seen by this model during pretraining], demonstrates this effectiveness.

\begin{figure*}[h] 
    \centering
    \includegraphics[width=1\linewidth]{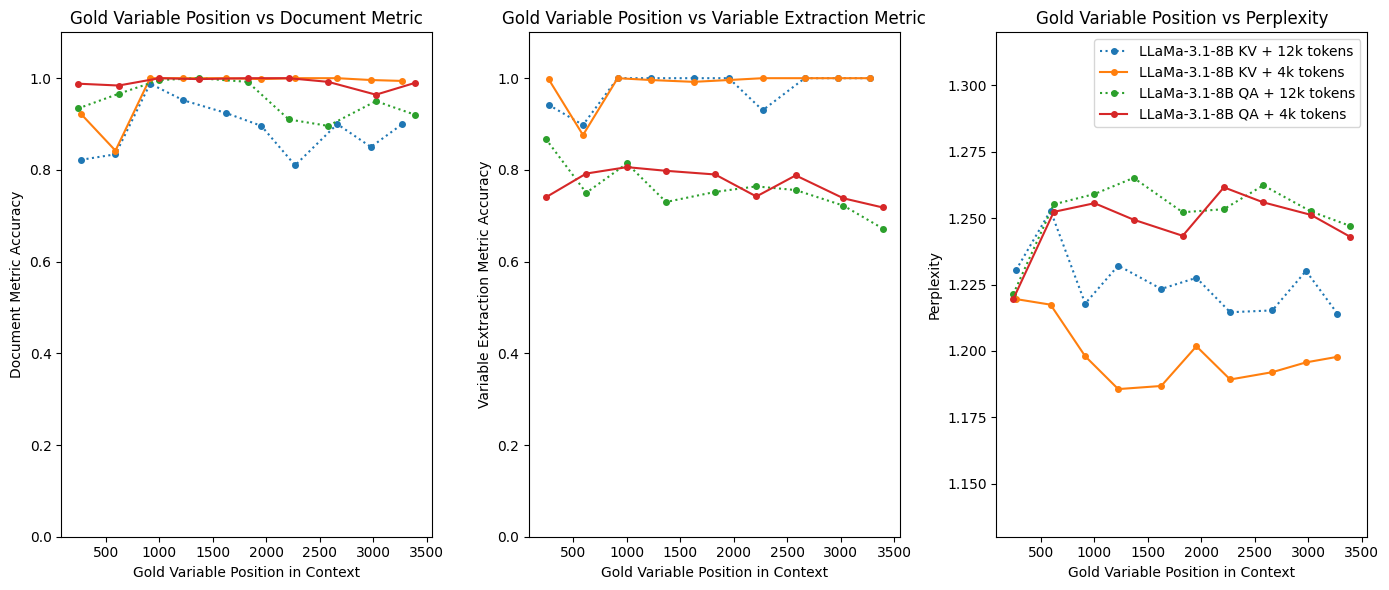} 
    \caption{Performance of LLaMa-3.1-8B across Tasks for 4k \& 12k tokens}
    \label{fig:llama3.1-across-tasks}
\end{figure*}

As illustrated in Figure \ref{fig:llama3.1-across-tasks}, the LLaMa-3.1-8B model exhibits near-perfect performance across both the Key-Value (KV) and Question-Answering (QA) tasks when constrained to 4k tokens. This finding validates the hypothesis that \textbf{long-context pre-training inherently grants superior positional robustness at shorter sequence lengths}.

However, Figure \ref{fig:llama3.1-across-tasks} also reveals two critical behaviors:
\begin{enumerate}
    \item \textbf{Performance Degradation with Length:} Performance noticeably decreases at the same document positions when the context length is increased to 12k tokens, a trend that is explored further in the subsequent section. This demonstrates that while long-context models are superior, the positional bias remains a function of the relative fullness of the context window.
    \item \textbf{Persistent Data Bias:} The model exhibits clear bias across tasks. The QA format consistently results in lower accuracy compared to the KV format, a finding quantitatively supported by the corresponding increase in perplexity. This confirms that even the most advanced models are susceptible to data representation bias.
\end{enumerate}

Crucially, the high performance metrics reported by LLaMa-3.1-8B in Table \ref{table:doc-metric-performance} and Table \ref{table:var-metric-performance} establish that this task is demonstrably solvable by models of just 8 billion parameters. This strongly suggests that while model parameters are a factor in overall capability, the "lost-in-the-middle" phenomenon is primarily governed by the model's pre-training sequence length rather than its sheer size.

\subsection{Degradation of Long-Context Model Performance with Increased Context Length}
\label{subsec:degradation}
Analysis of the results presented in Table \ref{table:long-context-performance} confirms that performance for both LLaMa-3.1-8B and FILM-7B is superior at 4k tokens compared to 12k tokens, establishing that positional robustness remains relative to the context window's utilization.

Our findings reveal a critical difference in the degradation across metrics. The Document Metric-which assesses the model's spatial awareness-experiences a significantly more severe decline than the Variable Extraction Metric. Specifically, on the Key-Value task, LLaMa-3.1-8B's performance declined by $12\%$, while FILM-7B saw a steep reduction of $31\%$. Similarly, the Question-Answering task saw drops of $10\%$ for both models. In contrast, the Variable Extraction Metric, which measures semantic retrieval, declined by only $5\%$ on average across all tasks and models.

\begin{figure*}[h] 
    \centering
    \includegraphics[width=1\linewidth]{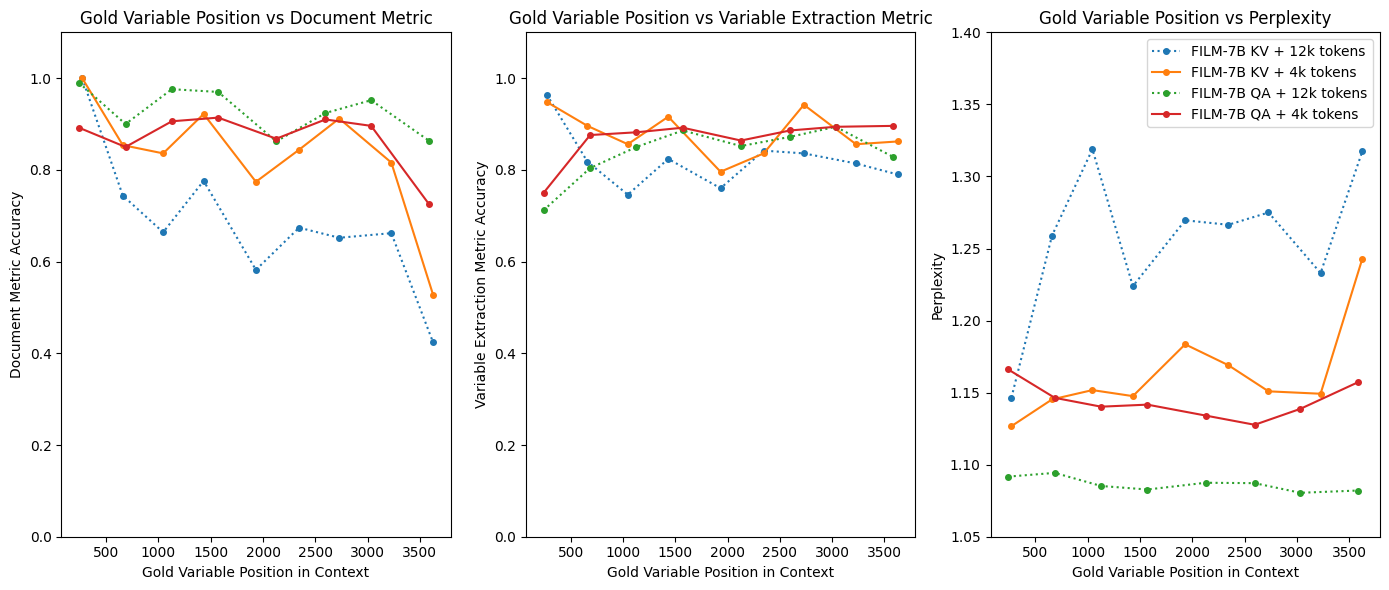} 
    \caption{Performance of FILM-7B across Tasks for 4k \& 12k tokens}
    \label{fig:film7b-across-tasks}
\end{figure*}

\begin{figure*}[h] 
    \centering
    \includegraphics[width=1\linewidth]{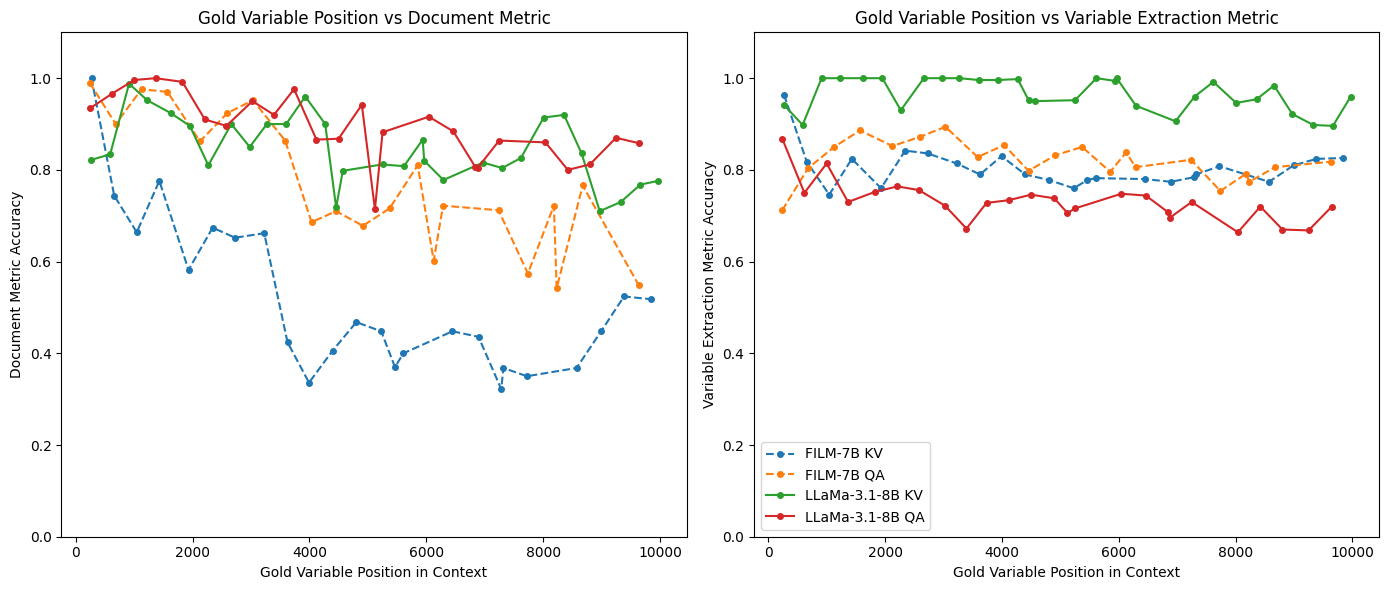} 
    \caption{Performance of Long Context Models Across Tasks on 10k tokens}
    \label{fig:long-context-performance}
\end{figure*}

This differential degradation leads to a key inference: as context length increases, the model's ability to determine \textbf{where} the target information is located (Document Metric) degrades significantly, even affecting tokens at the periphery of the context. However, its ability to understand \textbf{what} the answer is (Variable Extraction Metric) remains comparatively stable. This indicates that while long-context models maintain semantic comprehension, their internal spatial mapping of the context is highly fragile under increasing density. This behavior is visually correlated by comparing the initial 4k-token performance shown in Figure \ref{fig:llama3.1-across-tasks} and Figure \ref{fig:film7b-across-tasks}, with the full 10k-token performance illustrated in Figure \ref{fig:long-context-performance}.

\subsection{Limitations}
\label{subsec:limitations}
Our experimental scope utilized a benchmark effective up to approximately 12k tokens, which was sufficient for evaluating current- and previous-generation models. However, the rapidly evolving landscape of LLMs now includes models boasting context windows exceeding 500k tokens and even reaching the 1-2 million token range \cite{comanici2025gemini, rando2025longcodebench}. While existing benchmarks like RULER provide valuable context, future work is necessary to expand our GM-Extract benchmark to these multi-hundred-thousand token scales to rigorously test the persistence and nature of the "lost-in-the-middle" problem in extreme long-context settings.

Due to inherent institutional and resource constraints, our study on mitigation focused exclusively on fine-tuning effects, demonstrating that methods like PEFT/LoRA can only alleviate the positional bias. Our results consistently suggest that the root cause of context generalization issues lies in the pre-training stage. Therefore, comprehensive studies are necessary to evaluate the fundamental effects of both the pre-training objective and the subsequent Reinforcement Learning from Human Feedback (RLHF) stages on a model's long-context capabilities.

\subsection{Future Research Questions}
Based on our findings, we propose the following research agenda for the field:
\paragraph{RQ1} Given the observed differential degradation where the Document Metric (spatial awareness) collapses far more severely than the Variable Extraction Metric (semantic comprehension), how can a dedicated, normalized spatial awareness metric be integrated into standard evaluation suites to provide a clearer diagnostic signal of long-context failure modes?
\paragraph{RQ2} To what extent does modifying the pre-training sequence distribution or the curriculum of the RLHF phase directly mitigate the "lost-in-the-middle" problem and the observed data representation bias, and how do these effects compare against white-box and black-box inference-time interventions?
\paragraph{RQ3} What are the inherent limits of data-driven mitigation strategies (e.g., IN2-Training) in addressing positional bias for models with extremely long context windows (e.g., $>100k$ tokens), and does the required volume of augmented training data scale linearly with the context length to maintain positional robustness?

\section{Conclusion}
\label{sec:conclusion}
In this work, we conducted a rigorous investigation into the \textbf{"lost-in-the-middle" problem} within a practical context, addressing the critical reliability concerns of deploying Large Language Models (LLMs) in real-world information retrieval applications. We introduced \textbf{GM-Extract}, a novel benchmark dataset, alongside two distinct evaluation metrics-the \textbf{Document Metric} and the \textbf{Variable Extraction Metric}-to precisely diagnose specific failure modes in context utilization.

Our empirical findings established three decisive conclusions about the behavior of LLMs in long-context retrieval:
\begin{enumerate}
    \item We demonstrated that model performance is significantly governed by the \textbf{format of the input data}, even when the content is semantically identical, confirming a strong bias inherited from pre-training.
    \item We confirmed that the performance ceiling for short-context tasks (4k tokens) is decisively raised by models pre-trained on much longer sequence lengths (e.g., LLaMa-3.1-8B), establishing that the ability to effectively utilize context is primarily a function of the model's \textbf{pre-training context length}.
    \item We showed that as context density increases, the models \textbf{retain semantic comprehension} ("what" the answer is) but critically \textbf{lose their spatial awareness} ("where" the answer is located).
\end{enumerate}
Finally, our evaluation of mitigation strategies revealed the efficacy and limitations of both white-box (MS-PoE, Hidden State Scaling) and black-box (Information-Intensive Training with PEFT) approaches, showing that optimal results often require combining an instruction-tuned base model with targeted positional data augmentation. Our findings underscore that LLMs inherit minute data biases at all stages, necessitating that future research focus equally on the impact of pre-training context generalization and the robustness to varied data representations.

\section{Acknowledgments}
\label{sec:acknowledgement}
This work represents the culmination of several months of dedicated effort and collaboration. We sincerely thank all individuals who provided invaluable technical and intellectual support, particularly those whose critical involvement was instrumental in bringing this paper to fruition.

We extend our deepest gratitude to the authors who provided their exceptional support. In particular, we acknowledge one co-author for providing the initial mentorship that transformed a desire to pursue research into a focused, executable study. We also recognize the significant intellectual contribution of the second co-author, whose challenging of our initial concepts ultimately led to the genesis and successful rigor of this work. Their continuous guidance and perseverance, especially in the face of various external roadblocks, proved essential to the quality and completion of this manuscript.

\bibliographystyle{unsrtnat}
\bibliography{template}

@article{liu2024lost,
  title={Lost in the Middle: How Language Models Use Long Contexts},
  author={Liu, Nelson F and Lin, Kevin and Hewitt, John and Paranjape, Ashwin and Bevilacqua, Michele and Petroni, Fabio and Liang, Percy},
  journal={Transactions of the Association for Computational Linguistics},
  volume={12},
  year={2024}
}

@article{bedi2024systematic,
  title={A systematic review of testing and evaluation of healthcare applications of large language models (LLMs)},
  author={Bedi, Suhana and Liu, Yutong and Orr-Ewing, Lucy and Dash, Dev and Koyejo, Sanmi and Callahan, Alison and Fries, Jason A and Wornow, Michael and Swaminathan, Akshay and Lehmann, Lisa Soleymani and others},
  journal={medRxiv},
  pages={2024--04},
  year={2024},
  publisher={Cold Spring Harbor Laboratory Press}
}

@inproceedings{liuagentbench,
  title={AgentBench: Evaluating LLMs as Agents},
  author={Liu, Xiao and Yu, Hao and Zhang, Hanchen and Xu, Yifan and Lei, Xuanyu and Lai, Hanyu and Gu, Yu and Ding, Hangliang and Men, Kaiwen and Yang, Kejuan and others},
  booktitle={The Twelfth International Conference on Learning Representations}
}

@article{pawar2024and,
  title={The What, Why, and How of Context Length Extension Techniques in Large Language Models--A Detailed Survey},
  author={Pawar, Saurav and Tonmoy, SM and Zaman, SM and Jain, Vinija and Chadha, Aman and Das, Amitava},
  journal={arXiv preprint arXiv:2401.07872},
  year={2024}
}

@article{wang2025automating,
  title={Automating a complete software test process using LLMs: An automotive case study},
  author={Wang, Shuai and Yu, Yinan and Feldt, Robert and Parthasarathy, Dhasarathy},
  journal={arXiv preprint arXiv:2502.04008},
  year={2025}
}

@article{touvron2023llama,
  title={Llama 2: Open foundation and fine-tuned chat models},
  author={Touvron, Hugo and Martin, Louis and Stone, Kevin and Albert, Peter and Almahairi, Amjad and Babaei, Yasmine and Bashlykov, Nikolay and Batra, Soumya and Bhargava, Prajjwal and Bhosale, Shruti and others},
  journal={arXiv preprint arXiv:2307.09288},
  year={2023}
}

@misc{chiang2023vicuna,
  title        = {Vicuna: An open-source chatbot impressing GPT-4 with 90\%* ChatGPT quality},
  author       = {Wei-Lin Chiang and Zhuohan Li and Zi Lin and Ying Sheng and Zhanghao Wu and Hao Zhang and Lianmin Zheng and Siyuan Zhuang and Yonghao Zhuang and Joseph E. Gonzalez and Ion Stoica},
  year         = {2023},
  howpublished = {\url{https://vicuna.lmsys.org}},
  note         = {Accessed: 2023-04-14}
}

@article{grattafiori2024llama,
  title={The llama 3 herd of models},
  author={Grattafiori, Aaron and Dubey, Abhimanyu and Jauhri, Abhinav and Pandey, Abhinav and Kadian, Abhishek and Al-Dahle, Ahmad and Letman, Aiesha and Mathur, Akhil and Schelten, Alan and Vaughan, Alex and others},
  journal={arXiv preprint arXiv:2407.21783},
  year={2024}
}

@article{hsieh2024ruler,
  title={RULER: What's the Real Context Size of Your Long-Context Language Models?},
  author={Hsieh, Cheng-Ping and Sun, Simeng and Kriman, Samuel and Acharya, Shantanu and Rekesh, Dima and Jia, Fei and Zhang, Yang and Ginsburg, Boris},
  journal={arXiv preprint arXiv:2404.06654},
  year={2024}
}

@article{bai2023longbench,
  title={Longbench: A bilingual, multitask benchmark for long context understanding},
  author={Bai, Yushi and Lv, Xin and Zhang, Jiajie and Lyu, Hongchang and Tang, Jiankai and Huang, Zhidian and Du, Zhengxiao and Liu, Xiao and Zeng, Aohan and Hou, Lei and others},
  journal={arXiv preprint arXiv:2308.14508},
  year={2023}
}

@article{shaham2022scrolls,
  title={Scrolls: Standardized comparison over long language sequences},
  author={Shaham, Uri and Segal, Elad and Ivgi, Maor and Efrat, Avia and Yoran, Ori and Haviv, Adi and Gupta, Ankit and Xiong, Wenhan and Geva, Mor and Berant, Jonathan and others},
  journal={arXiv preprint arXiv:2201.03533},
  year={2022}
}

@inproceedings{hsieh2024found,
  title={Found in the middle: Calibrating Positional Attention Bias Improves Long Context Utilization},
  author={Hsieh, Cheng-Yu and Chuang, Yung-Sung and Li, Chun-Liang and Wang, Zifeng and Le, Long and Kumar, Abhishek and Glass, James and Ratner, Alexander and Lee, Chen-Yu and Krishna, Ranjay and others},
  booktitle={Findings of the Association for Computational Linguistics ACL 2024},
  pages={14982--14995},
  year={2024}
}

@article{zhang2025lost,
  title={Lost-in-the-Middle in Long-Text Generation: Synthetic Dataset, Evaluation Framework, and Mitigation},
  author={Zhang, Junhao and Zhang, Richong and Kong, Fanshuang and Miao, Ziyang and Ye, Yanhan and Zheng, Yaowei},
  journal={arXiv preprint arXiv:2503.06868},
  year={2025}
}

@misc{kaiokendev_context_2023,
  author       = {kaiokendev},
  title        = {Extending Context is Hard … but not Impossible},
  howpublished = {\url{https://kaiokendev.github.io/context}},
  year         = {2023},
  note         = {Accessed: 2025-10-01}
}

@article{chen2023extending,
  title={Extending context window of large language models via positional interpolation},
  author={Chen, Shouyuan and Wong, Sherman and Chen, Liangjian and Tian, Yuandong},
  journal={arXiv preprint arXiv:2306.15595},
  year={2023}
}

@article{su2024roformer,
  title={RoFormer: Enhanced transformer with Rotary Position Embedding},
  author={Su, Jianlin and Ahmed, Murtadha and Lu, Yu and Pan, Shengfeng and Bo, Wen and Liu, Yunfeng},
  journal={Neurocomputing},
  volume={568},
  pages={127063},
  year={2024},
  publisher={Elsevier}
}

@article{peng2023yarn,
  title={Yarn: Efficient context window extension of large language models},
  author={Peng, Bowen and Quesnelle, Jeffrey and Fan, Honglu and Shippole, Enrico},
  journal={arXiv preprint arXiv:2309.00071},
  year={2023}
}

@article{zhang2024found,
  title={Found in the middle: How language models use long contexts better via plug-and-play positional encoding},
  author={Zhang, Zhenyu and Chen, Runjin and Liu, Shiwei and Yao, Zhewei and Ruwase, Olatunji and Chen, Beidi and Wu, Xiaoxia and Wang, Zhangyang and others},
  journal={Advances in Neural Information Processing Systems},
  volume={37},
  pages={60755--60775},
  year={2024}
}

@article{yu2024mitigate,
  title={Mitigate position bias in large language models via scaling a single dimension},
  author={Yu, Yijiong and Jiang, Huiqiang and Luo, Xufang and Wu, Qianhui and Lin, Chin-Yew and Li, Dongsheng and Yang, Yuqing and Huang, Yongfeng and Qiu, Lili},
  journal={arXiv preprint arXiv:2406.02536},
  year={2024}
}

@article{an2024make,
  title={Make your llm fully utilize the context},
  author={An, Shengnan and Ma, Zexiong and Lin, Zeqi and Zheng, Nanning and Lou, Jian-Guang and Chen, Weizhu},
  journal={Advances in Neural Information Processing Systems},
  volume={37},
  pages={62160--62188},
  year={2024}
}

@article{hu2022lora,
  title={Lora: Low-rank adaptation of large language models.},
  author={Hu, Edward J and Shen, Yelong and Wallis, Phillip and Allen-Zhu, Zeyuan and Li, Yuanzhi and Wang, Shean and Wang, Lu and Chen, Weizhu and others},
  journal={ICLR},
  volume={1},
  number={2},
  pages={3},
  year={2022}
}

@article{xiao2023efficient,
  title={Efficient streaming language models with attention sinks},
  author={Xiao, Guangxuan and Tian, Yuandong and Chen, Beidi and Han, Song and Lewis, Mike},
  journal={arXiv preprint arXiv:2309.17453},
  year={2023}
}

@inproceedings{kandpal2023large,
  title={Large language models struggle to learn long-tail knowledge},
  author={Kandpal, Nikhil and Deng, Haikang and Roberts, Adam and Wallace, Eric and Raffel, Colin},
  booktitle={International conference on machine learning},
  pages={15696--15707},
  year={2023},
  organization={PMLR}
}

@article{mallen2022not,
  title={When not to trust language models: Investigating effectiveness of parametric and non-parametric memories},
  author={Mallen, Alex and Asai, Akari and Zhong, Victor and Das, Rajarshi and Khashabi, Daniel and Hajishirzi, Hannaneh},
  journal={arXiv preprint arXiv:2212.10511},
  year={2022}
}

@misc{kamradt2023needle,
  author       = {Gregory Kamradt},
  title        = {Needle In A Haystack - Pressure Testing LLMs},
  howpublished = {\url{https://github.com/gkamradt/LLMTestNeedleInAHaystack/tree/main}},
  year         = {2023}
}

@article{comanici2025gemini,
  title={Gemini 2.5: Pushing the frontier with advanced reasoning, multimodality, long context, and next generation agentic capabilities},
  author={Comanici, Gheorghe and Bieber, Eric and Schaekermann, Mike and Pasupat, Ice and Sachdeva, Noveen and Dhillon, Inderjit and Blistein, Marcel and Ram, Ori and Zhang, Dan and Rosen, Evan and others},
  journal={arXiv preprint arXiv:2507.06261},
  year={2025}
}

@article{rando2025longcodebench,
  title={LongCodeBench: Evaluating Coding LLMs at 1M Context Windows},
  author={Rando, Stefano and Romani, Luca and Sampieri, Alessio and Franco, Luca and Yang, John and Kyuragi, Yuta and Galasso, Fabio and Hashimoto, Tatsunori},
  journal={arXiv preprint arXiv:2505.07897},
  year={2025}
}

@article{peysakhovich2023attention,
  title={Attention sorting combats recency bias in long context language models},
  author={Peysakhovich, Alexander and Lerer, Adam},
  journal={arXiv preprint arXiv:2310.01427},
  year={2023}
}

@article{chen2023fortify,
  title={Fortify the shortest stave in attention: Enhancing context awareness of large language models for effective tool use},
  author={Chen, Yuhan and Lv, Ang and Lin, Ting-En and Chen, Changyu and Wu, Yuchuan and Huang, Fei and Li, Yongbin and Yan, Rui},
  journal={arXiv preprint arXiv:2312.04455},
  year={2023}
}

@article{chen2023clex,
  title={Clex: Continuous length extrapolation for large language models},
  author={Chen, Guanzheng and Li, Xin and Meng, Zaiqiao and Liang, Shangsong and Bing, Lidong},
  journal={arXiv preprint arXiv:2310.16450},
  year={2023}
}

@article{jin2024llm,
  title={Llm maybe longlm: Self-extend llm context window without tuning},
  author={Jin, Hongye and Han, Xiaotian and Yang, Jingfeng and Jiang, Zhimeng and Liu, Zirui and Chang, Chia-Yuan and Chen, Huiyuan and Hu, Xia},
  journal={arXiv preprint arXiv:2401.01325},
  year={2024}
}

@article{press2021train,
  title={Train short, test long: Attention with linear biases enables input length extrapolation},
  author={Press, Ofir and Smith, Noah A and Lewis, Mike},
  journal={arXiv preprint arXiv:2108.12409},
  year={2021}
}

@article{ding2024longrope,
  title={Longrope: Extending llm context window beyond 2 million tokens},
  author={Ding, Yiran and Zhang, Li Lyna and Zhang, Chengruidong and Xu, Yuanyuan and Shang, Ning and Xu, Jiahang and Yang, Fan and Yang, Mao},
  journal={arXiv preprint arXiv:2402.13753},
  year={2024}
}

@article{mohtashami2023random,
  title={Random-access infinite context length for transformers},
  author={Mohtashami, Amirkeivan and Jaggi, Martin},
  journal={Advances in Neural Information Processing Systems},
  volume={36},
  pages={54567--54585},
  year={2023}
}
\clearpage

\begin{appendices}
\section{Fine-tuning Ablation Results for QA Task on Vicuna-7B-1.5}
\label{app:fine-tuning}
To determine the optimal configuration for our data-driven mitigation strategy, we conducted an ablation study on the instruction-tuned Vicuna-7B-1.5 model for the Question-Answering (QA) task. The ablation focused on three primary experimental variables:

\begin{enumerate}
    \item \textbf{LoRA Hyperparameters:} The LoRA rank ($\textit{r}$) and scaling coefficient ($\alpha$) were held in constant ratio across all experiments, while we do change the parameters to understand it's effect on finetuning.
    \item \textbf{Total Data Volume:} We experimented with three distinct total volumes of benchmark permutations: 60k, 100k, and 120k samples.
    \item \textbf{Positional Data Augmentation (MIX):} This non-uniform distribution strategy utilized $50\%$ of the data generated with a uniform positional distribution and $50\%$ of the data generated exclusively on the middle document positions, where the model is known to struggle.
\end{enumerate}

\begin{figure}[h] 
    \centering
    \includegraphics[width=0.75\linewidth]{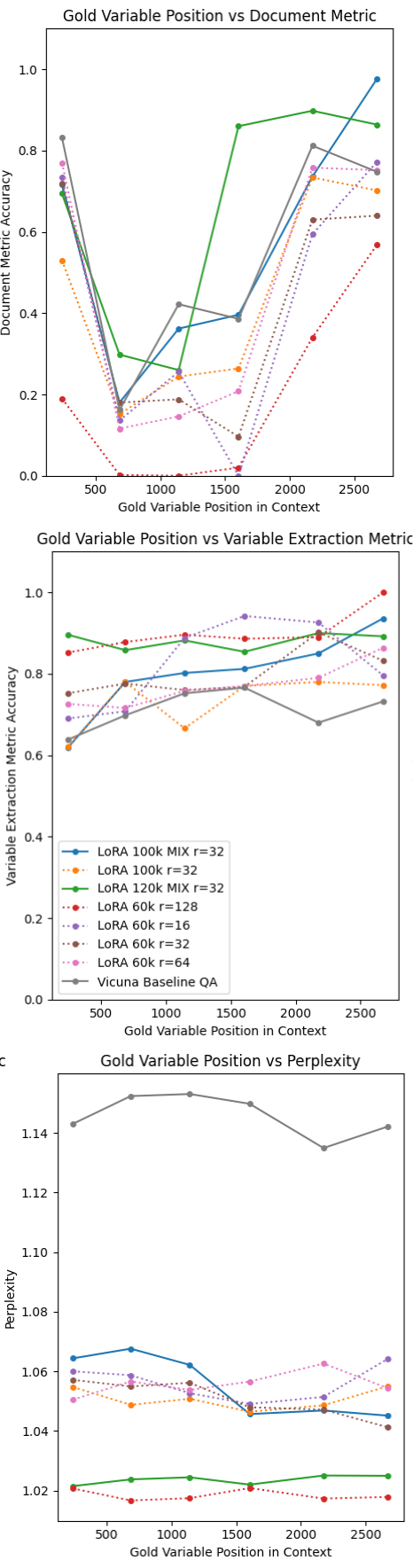} 
    \caption{Finetuning Ablation results }
    \label{fig:finetuning-qa-results}
\end{figure}

Figure \ref{fig:finetuning-qa-results} compares the performance of the various fine-tuning configurations against the baseline Vicuna-7B-1.5 model. The results clearly indicate that the most significant performance gain in the challenging middle region is achieved by the Positional Data Augmentation (MIX) strategy. This finding validates our hypothesis that explicitly over-sampling data in failure-prone context regions is an effective technique for alleviating positional bias.

\section{List of Questions}
\label{app:list-questions}
\begin{enumerate}
    \item What input does the variable ENUM\_AIR\_CtrlMode refer to?
    \item What input does the variable DARRAY\_b\_SgnlRatnlztnsFail[CIDX\_i\_12p5msIdx] refer to?
    \item What is the Data Type of the Variable VAR\_BatSysBatterySOCError\_rsp ?
    \item What is the Data Type of the Variable DUAL\_e\_SysCurrentState ?
    \item How do you SET the Variable VAR\_BatSysBatteryCurrentExtendedRange\_rsp\_Inv ?
    \item How do you SET the Variable VARRAY\_b\_PwrSupPerfFA ?
    \item What is the Storage Type of the Variable DUAL\_e\_SafetySysSt ?
    \item What is the Storage Type of the Variable VAR\_b\_NonPrimSnsrVsMtrFA ?
    \item What is the Quantity Measured by the Variable DUAL\_T\_SysCompTempFiltd ?
    \item What is the Quantity Measured by the Variable VAR\_BatHeatPwrCmd ?
\end{enumerate}

\clearpage
\section{Prompts Utilized for Tasks}
\label{app:prompts}
\lstset{
  basicstyle=\ttfamily\small,
  breaklines=true,
  frame=single,
  backgroundcolor=\color{gray!10},
  keywordstyle=\color{blue},
  numbers=none,
  xleftmargin=0pt,
  framexleftmargin=0pt,
  literate={\\land}{{$\land$}}1
           {\\lnot}{{$\lnot$}}1
           {\\neq}{{$\neq$}}1
}
\subsection{Key-Value Task}
\begin{lstlisting}[language=]
You will be given a long document containing key-value pairs. Your task is to find the exact value associated with the key provided in the question at the end.
Given the data of variables in the context you will be given a name of the variable at the end given by Variable, return ONLY THE:
1. ID of the Document referred 
2. {to_retrieve}

Key-Value Variable Data:
{kv_data}

Variable: "{variable}"
What is the {to_retrieve} and Document ID for the variable above?
Document ID | {to_retrieve}:
\end{lstlisting}
\subsection{Question-Answer Task}
\begin{lstlisting}[language=]
You will be given a long document containing key-value pairs. Your task is to provide a short high-quality answer for the question provided by Question tag based on the key value data:

Key-Value Variable Data:
{kv_data}

Question: "{question} Also provide the ID of the Document referred (IMPORTANT)."
Answer:
\end{lstlisting}
\end{appendices}
\end{document}